%% file: main.tex
\documentclass{MITcsail}

\input{tex/command/math_commands}

\input{tex/command/mlVecMat}

\input{tex/command/problem_blocks}
\usepackage{tikz}
\usepackage{pgfplots}
\pgfplotsset{compat=1.18}
\usepackage{pgf-pie}
\usepackage{listings}
\usepackage[dvipsnames]{xcolor} 
\usepackage{xspace}
\usepackage{inconsolata}

\newcommand{\benchmark}{FrontierCS\xspace}

\newcommand{\pagehead}{FrontierCS}
\usepackage{authblk}

\newcommand{\cpt}{\fontsize{18.5pt}{20pt}\selectfont}

\title{\textcolor[HTML]{002676}{\cpt FrontierCS: Evolving Challenges for Evolving Intelligence}}

\author{%
  \begin{minipage}{\textwidth}
    \raggedright
    \textbf{\small Contributors} \small ( $^*$equal contribution )\\ 
    \vspace{1ex}
    {\small
    \small
    Qiuyang Mang$^{1,*}$, Wenhao Chai$^{2,*}$, Zhifei Li$^{1, *}$, Huanzhi Mao$^{1,*}$, Shang Zhou$^{3,*}$, Alexander Du$^{1,4,*}$, \\
    \vspace{1ex}
    \small
    Hanchen Li$^{1,*}$, Shu Liu$^{1,*}$, Edwin Chen$^{5}$,  Yichuan Wang$^{1}$, Xieting Chu$^{6}$, Zerui Cheng$^{2}$, Yuan Xu$^{4}$, Tian Xia$^{1}$,}
    \vspace{1ex}

    {\small
     Zirui Wang$^{1}$, Tianneng Shi$^{1}$, Jianzhu Yao$^{2}$, Yilong Zhao$^{1}$, Qizheng Zhang$^{7}$, Charlie Ruan$^{1}$, Zeyu Shen$^{2}$,   \\
     \vspace{1ex}
     \small
     Kaiyuan Liu$^{8}$, Runyuan He$^{1}$, Dong Xing$^{4}$, Zerui Li$^{4}$,  Zirong Zeng$^{1}$, Yige Jiang$^{9}$, Lufeng Cheng$^{10}$, Ziyi Zhao$^{11}$,   \\
     \vspace{1ex}
     \small
     Youran Sun$^{1}$, Wesley Zheng$^{1}$, Meiyuwang Zhang$^{5}$, Ruyi Ji$^{12}$, Xuechang Tu$^{6}$, Zihan Zheng$^{13}$, Zexing Chen$^{3}$,   \\
     \vspace{1ex}
     \small
     Kangyang Zhou$^{14}$, Zhaozi Wang$^{13}$, Jingbang Chen$^{5}$
     }

    \vspace{1ex}

    \vspace{2ex}
    \textbf{\small Advisors} \small \textbf{( $^*$equal advising )}\\
    \vspace{1ex}
    \small
    Aleksandra Korolova$^{2}$, Peter Henderson$^{2}$, Pramod Viswanath$^{2}$, Vijay Ganesh$^{6}$, Saining Xie$^{13}$, Zhuang Liu$^{2}$, \\
    \vspace{1ex}
    \small
    Dawn Song$^{1}$, Sewon Min$^{1}$, Ion Stoica$^{1}$, Joseph E. Gonzalez$^{1,*}$, Jingbo Shang$^{3,*}$, Alvin Cheung$^{1,*}$
    \vspace{2ex}

    \textbf{\small Affiliations}\\
    \vspace{1ex}
    {\small $^{1}$UC Berkeley \quad $^{2}$Princeton University \quad $^{3}$UCSD \quad $^{4}$X-camp Academy \quad $^{5}$Independent \quad $^{6}$Georgia Tech\\
    \small \vspace{0.5ex} $^{7}$Stanford University \quad $^{8}$University of Washington \quad $^{9}$Nanyang Technological University \\ \small \vspace{0.5ex} $^{10}$University of Toronto \quad
    $^{11}$UIUC \quad $^{12}$University of Michigan \quad $^{13}$New York University \quad $^{14}$MIT \\}
  \end{minipage}%
}

\begin{document}

\maketitle
\thispagestyle{firstpagestyle} 

\input{tex/0_abs}
\clearpage
\input{tex/1_intro}
\input{tex/2_survey}
\input{tex/3_method}

\input{tex/4_exp}
\input{tex/4_examples}

\input{tex/5_conclusion}

\bibliographystyle{plain}
\bibliography{references}

\appendix

\end{document}

%% file: tex/command/math_commands.tex
\usepackage{amsmath,amsfonts,bm}

\def\eqref#1{equation~\ref{#1}}

\def\1{\bm{1}}

\DeclareMathAlphabet{\mathsfit}{\encodingdefault}{\sfdefault}{m}{sl}
\SetMathAlphabet{\mathsfit}{bold}{\encodingdefault}{\sfdefault}{bx}{n}



%% file: tex/command/mlVecMat.tex
\definecolor{darkred}{RGB}{140, 21, 21}
\definecolor{lightgray}{gray}{0.7}
\definecolor{orange}{HTML}{F58025}
\definecolor{deepred}{rgb}{0.631,0.102,0.102}
\definecolor{amethyst}{rgb}{0.6, 0.4, 0.8}
\definecolor{darkgreen}{rgb}{0.3,0.7,0.3}
\definecolor{salmon}{RGB}{241, 150, 141}
\definecolor{mildyellow}{HTML}{FFF2CC}

\usepackage{xcolor}
\usepackage[normalem]{ulem}

\usepackage{hyperref}
\usepackage{xcolor}
\hypersetup{
    colorlinks=true,
    linkcolor=darkred,
    citecolor=lightgray,
    filecolor=darkred,
    urlcolor=darkred
}
\usepackage[numbers, sort&compress]{natbib}

\usepackage{amsmath}
\usepackage{amsfonts}
\usepackage{amssymb}
\usepackage{wrapfig}
\usepackage{subcaption}
\usepackage{multirow}
 \usepackage{mathtools} 

\usepackage{verbatim}

\usepackage{anyfontsize}

\usepackage{microtype}
\usepackage{graphicx}
\usepackage{booktabs} 
\usepackage{multirow}
\usepackage{amsmath,amssymb}
\usepackage{booktabs}
\usepackage{caption,subcaption}
\usepackage{svg}
\usepackage{authblk}

\definecolor{mygreen}{HTML}{3cb44b}
\definecolor{skyblue}{HTML}{beffff}
\definecolor{lightgreen}{HTML}{90ee90}

\usepackage{color, colortbl}

\definecolor{emerald}{rgb}{0.31, 0.78, 0.37}

\usepackage{datetime}
\newdateformat{ymd}{\the\year-\the\month-\the\day}

\usepackage{tcolorbox}
\usepackage{enumitem}
\definecolor{mygreen}{HTML}{3cb44b}
\colorlet{myyellow}{green!10!orange!90!}
\makeatletter

\usepackage{tikz}
\usetikzlibrary{arrows,shapes,snakes,automata,backgrounds,fit,petri}
\usepackage{adjustbox}

\newcommand{\RN}[1]{%
	\textup{\lowercase\expandafter{\it \romannumeral#1}}%
}
\usepackage{tabu}

\newcommand{\ie}[0]{\emph{i.e., }}

\newcommand{\eg}[0]{\emph{e.g., }}

\newcommand{\beq}{\vspace{0mm}\begin{equation}}
\newcommand{\eeq}{\vspace{0mm}\end{equation}}
\newcommand{\beqs}{\vspace{0mm}\begin{eqnarray}}
\newcommand{\eeqs}{\vspace{0mm}\end{eqnarray}}
\newcommand{\barr}{\begin{array}}
\newcommand{\earr}{\end{array}}

\usepackage{color, colortbl}
\definecolor{Gray}{gray}{0.93}

\usepackage{lipsum}

\usepackage{pifont}

\usepackage{makecell}

\usepackage{xcolor,amsmath}
\usepackage[ruled,vlined,linesnumbered]{algorithm2e}
\SetKwComment{Comment}{$\triangleright$ }{}
\SetKwInput{KwInput}{Input}
\SetKwInput{KwOutput}{Output}
\DontPrintSemicolon

\usepackage{xcolor}
\definecolor{mygreen}{HTML}{3cb44b}

\SetKwComment{Comment}{\color{green!50!black}\# }{}

\SetKwProg{Function}{def}{:}{}

\SetKwProg{For}{for}{:}{}
\SetKwProg{If}{if}{:}{}

%% file: tex/command/problem_blocks.tex
\usepackage{tcolorbox}
\tcbuselibrary{skins,breakable}

\definecolor{brandblue}{HTML}{002676} 
\definecolor{brandorange}{HTML}{FDB515} 

\tcbset{enhanced,sharp corners,boxrule=0.6pt}

\newtcolorbox{sampleproblem}[2][]{
  colback=white,
  colframe=brandblue,
  colbacktitle=brandblue!7,
  coltitle=brandblue,
  fonttitle=\bfseries\large,
  title={#2},
  left=10pt,right=10pt,top=8pt,bottom=10pt,
  #1
}

\newcommand{\pbsection}[1]{\par\vspace{0.25em}\textbf{#1.}\;}

\newcommand{\problabel}[1]{\phantomsection\label{#1}}
\newcommand{\probref}[1]{\hyperref[prob:#1]{Problem~#1}}

%% file: tex/0_abs.tex
\begin{abstract}

\textbf{Abstract:} We introduce FrontierCS, a benchmark of 156 open-ended problems across diverse areas of computer science, designed and reviewed by experts, including CS PhDs and top-tier competitive programming participants and problem setters. 
Unlike existing benchmarks that focus on tasks with known optimal solutions, FrontierCS targets problems where {\em the optimal solution is unknown, but the quality of a solution can be objectively evaluated.} Models solve these tasks by implementing executable programs rather than outputting a direct answer.
FrontierCS includes algorithmic problems, which are often NP-hard variants of competitive programming problems with objective partial scoring, and research problems with the same property.
For each problem we provide an expert reference solution and an automatic evaluator. 
Combining open-ended design, measurable progress, and expert curation, 
FrontierCS provides a benchmark at the frontier of computer-science difficulty. Empirically, we find that frontier reasoning models still lag far behind human experts on both the algorithmic and research tracks, that increasing reasoning budgets alone does not close this gap, and that models often over-optimize for generating merely workable code instead of discovering high-quality algorithms and system designs.

\vspace{2mm}
\textbf{Project page:} 
\href{www.frontier-cs.org}{www.frontier-cs.org}

\vspace{2mm}
\textbf{Code and Data:}
\href{https://github.com/FrontierCS/Frontier-CS}{Frontier-CS GitHub}


\end{abstract}

%% file: tex/1_intro.tex
\section{Introduction}

The rapid progress of large language models (LLMs) is evident on numerous code and reasoning benchmarks~\cite{mao2025bfclv4web, patil2025bfcl, jain2024livecodebench, zheng2025livecodebench, ouyang2025kernelbench, xu2025icpcevalprobingfrontiersllm, wang2025ojbenchcompetitionlevelcode, quan2025codeelobenchmarkingcompetitionlevelcode, chen2021evaluating, zhuo2025bigcodebenchbenchmarkingcodegeneration, liu2023repobenchbenchmarkingrepositorylevelcode, ding2023crosscodeevaldiversemultilingualbenchmark, liu2023codegeneratedchatgptreally, Li_2022, xia2025leetcodedatasettemporaldatasetrobust, yang2025elaboration, fang2024mathodysseybenchmarkingmathematicalproblemsolving}, which largely comprise closed-form tasks with a single optimal answer and a pass-or-fail criterion. Yet many frontier problems in computer science are intrinsically open-ended, requiring nuanced trade-offs among quality, efficiency, and robustness~\cite{Chen2025MLRBench, nie2025uqassessinglanguagemodels, fan-etal-2024-nphardeval, chen2025heurigymagenticbenchmarkllmcrafted, li2025optbenchevaluatingllmagent, ma2025swefficiencylanguagemodelsoptimize}. 
While recent work has begun using LLMs to address unsolved or open-ended CS problems~\cite{cheng2025barbarians, ma2025algorithm, novikov2025alphaevolve, agrawal2025gepa, mang2025automated}, these efforts naturally evaluate models only within their specific application domains or on a small number of representative cases, leaving the field without a comprehensive, cross-domain benchmark.

In this paper, we introduce FrontierCS, a coding benchmark that evaluates LLMs on solving open-ended computer science problems, where no known closed-form or deterministic optimal solution exists in practice.
Unlike math or reasoning benchmarks that require a direct answer (e.g., AIME), FrontierCS requires models to implement executable programs to solve the problem (e.g., LiveCodeBench).
We focus on tasks where a global optimum is either unknown or practically unattainable, yet any proposed solution can be deterministically checked for validity and assigned a score by an automatic evaluator~\cite{Li2025NPENGINE, imajuku2025ale, wang2025cybergym}. This design measures the ability of models to implement effective and efficient algorithms rather than algorithms that exhaustively solve the problem.
FrontierCS includes optimization tasks for which no known optimal solver exists in practical time,
each task includes an expert reference solution and a deterministic automatic evaluator to facilitate objective comparisons and ensure reproducibility. 
Each FrontierCS task can be solved by submitting executable code: the evaluator runs the program on generated instances and scores its outputs by task-specific metrics under resource limits (\emph{e.g.,} time and memory usage). 
The model is prompted with the problem specification (and any required I/O or API stubs) and must produce a self-contained solver program.

\begin{figure}[t]
    \centering
    \includegraphics[width=0.98\linewidth]{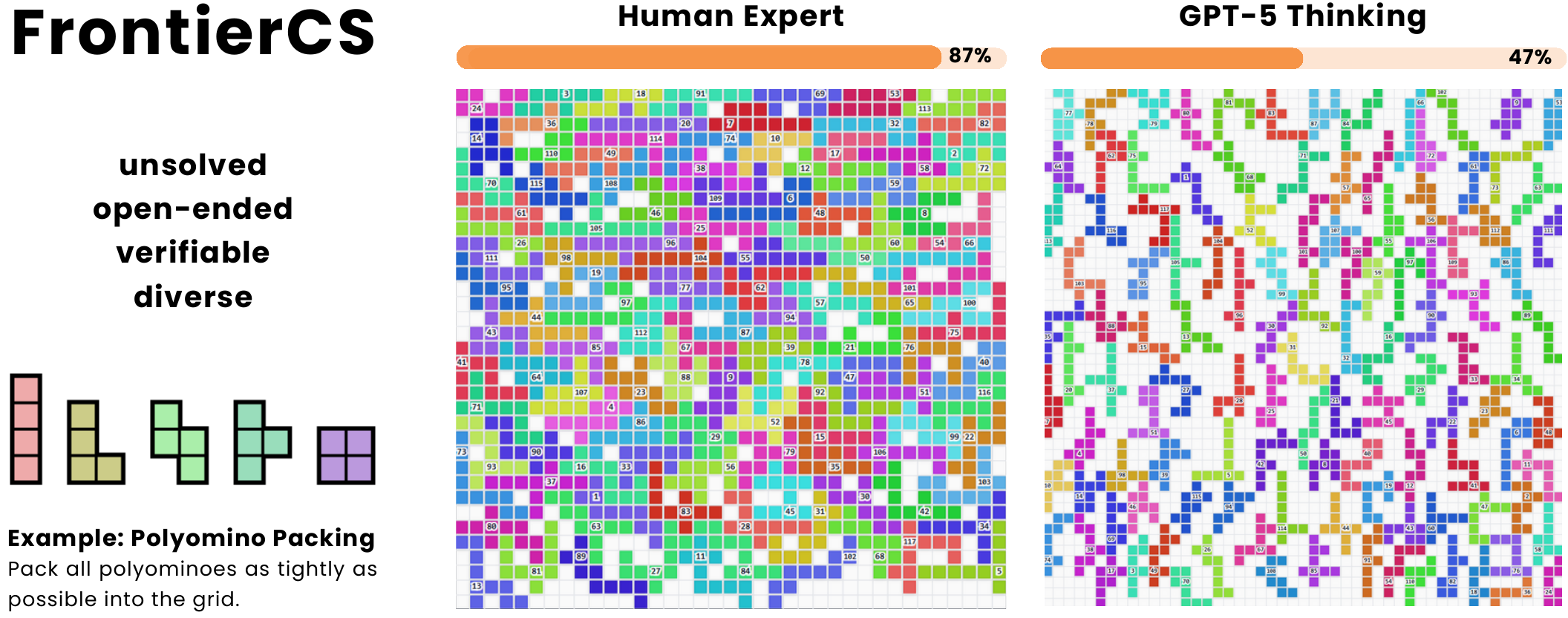}
    \caption{\textbf{FrontierCS}, an unsolved, open-ended, verifiable, and diverse benchmark for computer science tasks. The Polyomino Packing example, where both human experts and LLMs produce valid but non-optimal solutions that differ substantially in density. This reflects the benchmark's core design choice: problems are unsolved, admit many solution strategies, and are evaluated via deterministic scoring rather than pass-or-fail.}
    \label{fig:teaser}
\end{figure}

As an example of the type of tasks included in FrontierCS, consider the Polyomino Packing task in Figure~\ref{fig:teaser}, packing a set of $n$ polyominos (block shapes) into the smallest possible grid without overlapping to maximize density, \ie the area fraction occupied by the polyominoes.
Even though the optimal packing is unknown, solutions can still be compared objectively using packing density. The evaluator first runs the LLM-generated program to obtain a packing arrangement, then checks whether the arrangement is valid, \emph{i.e.,} if any polyominoes overlap or extend beyond the grid boundaries. If valid, the score is computed as the packing density, defined as the total area covered by all placed polyominoes divided by the area of the grid.
For this task, both the human expert and GPT-5 produce valid packings, but the human achieves 87\% density while GPT-5 achieves only 47\%. 
By varying $n$, this single specification yields infinitely many task instances, each with its own difficulty and unknown optimum.
This example illustrates the kind of tasks targeted by FrontierCS: problems with many valid solutions whose quality spans a continuous spectrum rather than a simple pass-or-fail outcome. 

Formally, we define an open-ended optimization problem without known polynomial time solutions as one that satisfies the following:
\begin{itemize}[leftmargin=*, itemsep=0.5pt]
\item \textbf{Unsolved or Intractable Optimum:} The global optimum is unknown to compute over all problem instances, requiring progress to come from creative algorithms, heuristics, search, or optimization.

\item \textbf{Deterministically Verifiable and Quantitatively Scored:} Solutions with runtime limit can be automatically checked for validity and assigned a numeric score that reflects the quality of the solution, rather than a simple pass-or-fail.

\item \textbf{Parametric Problem Generator:} The task specification induces a large, variable-difficulty space of instances, enabling fresh, unseen test cases to prevent leak and overfitting.

\end{itemize}

FrontierCS focuses on two tracks: algorithmic optimization problems, and tasks more closely tied to real-world CS research. Both tracks naturally exhibit open-endedness, stress-testing a model's ability to perform deep open-ended reasoning and discover nontrivial optimization strategies.
The design of FrontierCS encourages iterative improvement in an open-ended landscape rather than aiming for a deterministic optimal solution, since none of its problems have known practical optima.
FrontierCS, with its dynamic task scaling and objective quantitative feedback, offers an adaptive framework for continuous progress in LLM reasoning and creativity. Moreover, given that solutions are automatically verifiable and reward signals are available, FrontierCS is well-suited not only for evaluation but also for training and ablation studies. 
The scoring functions can effectively drive reinforcement learning or even self-play~\cite{zhou2025autocode, madaan2023selfrefineiterativerefinementselffeedback, chen2025spcevolvingselfplaycritic, fang2025serlselfplayreinforcementlearning, liu2025spiralselfplayzerosumgames, vanniekerk2025posttraininglargelanguagemodels, zweiger2025selfadaptinglanguagemodels, zhao2025absolute, wang2023selfinstructaligninglanguagemodels} as reward model. 

Empirically, we find that even the strongest frontier reasoning models remain far behind human experts on both the algorithmic and research tracks of FrontierCS. Simply scaling up context length or reasoning budgets yields diminishing returns on the hardest problems, and models frequently converge to locally workable but clearly suboptimal algorithms. These observations suggest that current large reasoning models are still missing key ingredients for truly open-ended computer-science problem solving, motivating FrontierCS as a challenging benchmark for future progress.

\begin{figure*}[t]
\centering

\newcommand{\AlgOpt}{29}
\newcommand{\AlgCons}{27}
\newcommand{\AlgInter}{51}
\newcommand{\AlgTotal}{107}

\newcommand{\ResOS}{8}
\newcommand{\ResHPC}{19}
\newcommand{\ResAI}{6}
\newcommand{\ResDB}{7}
\newcommand{\ResPL}{5}
\newcommand{\ResSec}{4}
\newcommand{\ResTotal}{49}

\resizebox{0.75\linewidth}{!}{%
\begin{tikzpicture}[font=\small]

  \begin{scope}[xshift=-2.2cm]
    \node[above=12pt, align=center] at (2,2.5)
      {Algorithmic Problems \\ (\textbf{Total \AlgTotal})};

    \begin{axis}[
      ybar,
      ymin=0,
      width=6cm,
      height=4cm,
      bar width=12pt,
      symbolic x coords={Optimization,Constructive,Interactive},
      xtick=data,
      xticklabel style={rotate=35, anchor=east},
      ylabel={\# Problems},
      enlarge x limits=0.25,
      nodes near coords,
      nodes near coords align={vertical},
      every node near coord/.append style={font=\scriptsize},
      axis x line*=bottom,
      axis y line*=left
    ]
      \addplot[fill=brandblue!60]
        coordinates {
          (Optimization,\AlgOpt)
          (Constructive,\AlgCons)
          (Interactive,\AlgInter)
        };
    \end{axis}
  \end{scope}

  \begin{scope}[xshift=4.2cm]
    \node[above=12pt, align=center] at (2,2.5)
      {Research Problems \\ (\textbf{Total \ResTotal})};

    \begin{axis}[
      ybar,
      ymin=0,
      width=6.5cm,
      height=4cm,
      bar width=10pt,
      symbolic x coords={OS,HPC,AI,DB,PL,Security},
      xtick=data,
      xticklabel style={rotate=35, anchor=east},
      ylabel={\# Problems},
      enlarge x limits=0.15,
      nodes near coords,
      nodes near coords align={vertical},
      every node near coord/.append style={font=\scriptsize},
      axis x line*=bottom,
      axis y line*=left
    ]
      \addplot[fill=brandorange!90]
        coordinates {
          (OS,\ResOS)
          (HPC,\ResHPC)
          (AI,\ResAI)
          (DB,\ResDB)
          (PL,\ResPL)
          (Security,\ResSec)
        };
    \end{axis}
  \end{scope}

\end{tikzpicture}%
}

\caption{\textbf{Categories} distribution of the 156 problems in FrontierCS. \textbf{Left}: Algorithmic Problems, covering Optimization tasks, Constructive tasks, and Interactive tasks, adapted from programming contests but rewritten into open-ended, partially scored variants. \textbf{Right}: Research Problems, spanning six major CS domains: OS (Operating Systems), HPC (High-Performance Computing), AI (Artificial Intelligence research tasks), DB (Databases), PL (Programming Languages), and Security (cybersecurity and vulnerability analysis). These research problems are sourced from real research workflows and each comes with a deterministic evaluator.
}
\label{fig:benchmark_overview}
\end{figure*}
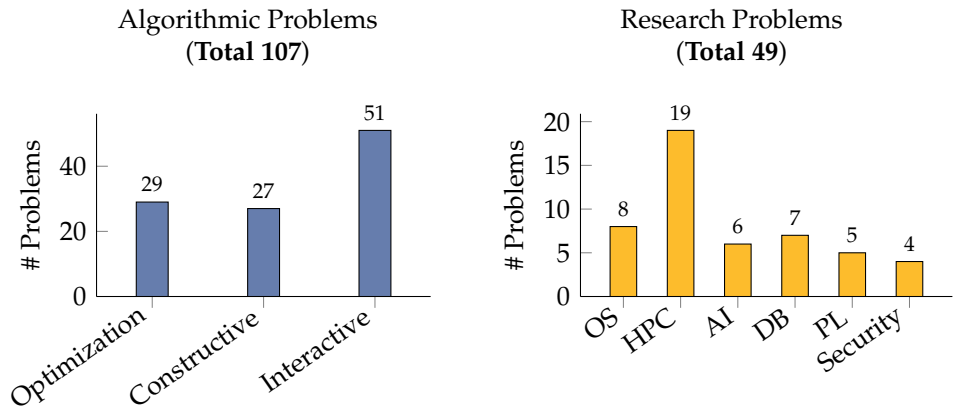

%% file: tex/2_survey.tex
\section{Related Work}

\paragraph{Closed-form benchmarks for code and reasoning.} A wealth of benchmarks have evaluated LLMs on coding and math problems with known solutions. In the programming domain, tasks like HumanEval~\cite{chen2021evaluating}, MBPP~\cite{austin2021program}, SWE-bench~\cite{jimenez2023swe}, BFCL~\cite{patil2025bfcl}, and LiveCodeBench~\cite{jain2024livecodebench} present coding challenges with unit tests as a pass-or-fail criterion. These were useful for early code models, but top LLMs now approach saturation on them. Similarly, in math and logical reasoning, datasets such as MATH~\cite{lightman2023let}, GSM8K~\cite{cobbe2021training}, and AIME were designed to test step-by-step reasoning. State-of-the-art models have achieved near-perfect scores on these high-school-level benchmarks, indicating that they no longer discriminate well at the frontier of ability. In the domain of mathematics, FrontierMath~\cite{glazer2024frontiermath} collects hundreds of new, expert-crafted math challenges spanning areas from algebraic geometry to number theory, each designed to require substantial creativity and insight. Crucially, every problem has a reference answer and an automated solution checker, allowing objective scoring even if the solution process is complex. Recently, LiveCodeBench Pro~\cite{zheng2025livecodebench} has increased the difficulty of its competitive programming benchmark and updates it quarterly. On the hardest split, only a few models are able to solve even a single problem. Even so, LiveCodeBench Pro still focuses only on problems that have known optimal solutions, whereas in many real-world scenarios, finding the optimal solution is tied to NP-completeness.

\paragraph{Open-ended and partially-scored benchmarks.} Beyond strictly one-answer tasks, a number of benchmarks embrace open-ended problem solving or partial-score evaluation. ALE-Bench~\cite{imajuku2025ale} introduces optimization-style tasks from AtCoder Heuristic Contests, replacing binary correctness with score-based evaluation. Spanning routing, packing, scheduling, and stochastic search, it challenges models to design heuristics that achieve high objective scores under strict runtime constraints. UQ~\cite{nie2025uqassessinglanguagemodels} curates a set of unsolved tasks from a collection of over 500 problems on forums like Stack Exchange, showing that even top models pass only around 15\% after human validation. MLR-Bench~\cite{Chen2025MLRBench} evaluates models on research tasks taken from top ML conferences, using an automated judge and agent. MLR-Bench finds that models are capable of writing coherent papers but still struggle with experimentation. Additionally, their automated judge aligns closely with human reviewers, revealing the possibility of automated evaluation for research tasks. NP-Engine~\cite{Li2025NPENGINE} introduces a benchmark and training framework for 10 classic NP-hard tasks, including Subset Sum, the Traveling Salesman Problem, and others. A model trained on their framework achieves state-of-the-art performance on these tasks and shows strong out-of-domain generalization.
KernelBench~\cite{ouyang2025kernelbench} evaluates LLMs on writing efficient GPU kernels for a variety of computational tasks, using performance metrics like correctness and runtime for scoring.

Existing benchmarks generally adopt one of three evaluation paradigms: binary pass-or-fail testing (\emph{e.g.,} LiveCodeBench), performance- or runtime-based scoring (\emph{e.g.,} KernelBench), or open-ended tasks that require human or LLM-as-judge evaluation (\emph{e.g.,} UQ).
Although ALE-Bench and the AtCoder Heuristic Contests fall outside these categories by using optimization-style scoring, they still cover a narrow slice of problem types, and their tasks are authored by only a small number of recurring problem setters, limiting their diversity.
FrontierCS addresses these gaps by providing a large, diverse, and continually expanding collection of real-world reasoning problems sourced from multiple platforms. Each problem is unsolved yet automatically verifiable, enabling objective evaluation without relying on binary correctness or human judgment.
Instead of checking only for optimality or execution performance, FrontierCS supports continuous scoring based on solution quality, allowing models to be credited for partial progress on research-style tasks where optimal solutions are unknown or computationally prohibitive.
Together, these properties establish FrontierCS as a new baseline for evaluating frontier model reasoning in open-ended, real-world problem settings.

%% file: tex/3_method.tex
\section{Problem Collection}

As shown in Figure~\ref{fig:benchmark_overview}, \benchmark consists 156 problems across two tracks: Algorithmic Problems and Research Problems. 
The Algorithmic Problems track contains 107 problems adapted from programming contests, covering three categories: Optimization, Constructive, and Interactive problems. 
The Research Problems track contains 49 problems sourced from real-world computer science research questions, spanning six domains: Operating Systems, High-Performance Computing, Artificial Intelligence, Databases, Programming Languages, and Security. 
Each problem is carefully curated through a multi-stage process involving proposal, implementation, and review to ensure quality and relevance, which will be detailed in the following sections.

\vspace{-2ex}
\paragraph{Taxonomy.}

To organize the algorithmic problems in \benchmark, we adopt a taxonomy that reflects the dominant reasoning mode each problem requires. 
\textbf{(1)} \emph{Constructive problems} center on synthesizing a valid structured object (\emph{e.g.,} graph and packing) under global constraints; even when tasks ask for a minimal or smallest solution, the difficulty lies in producing a coherent structure rather than tuning explicit parameters (\emph{e.g.,} \probref{1}, \probref{4} and \probref{5} in Section~\ref{sec:example}). 
\textbf{(2)} \emph{Optimization problems}, in contrast, require explicit search over a parameterized space to minimize or maximize a quantitative objective, often involving accuracy-latency or cost-capacity trade-offs (\emph{e.g.,} \probref{2} in Section~\ref{sec:example}). 
\textbf{(3)} \emph{Interactive problems} involve solving a hidden-instance task through a query-response protocol, where each action depends on previous replies, and performance typically depends not only on correctness but also on interaction efficiency such as minimizing the number of steps (\emph{e.g.,} \probref{3} and \probref{10} in Section~\ref{sec:example}). 
Together, these three categories capture the principal forms of algoritmic and open-ended reasoning that arise in real computer science tasks.

For problems in the Research Problems track, we categorize them based on their respective computer science domains, including Operating Systems (OS), High-Performance Computing (HPC), Artificial Intelligence (AI), Databases (DB), Programming Languages (PL), and Security. 
Specifically, when a research topic spans multiple domains, we assign it to the domain that the model's solution is primarily meant to address.
This domain-based taxonomy reflects the diverse challenges and methodologies inherent in different areas of computer science research.


\subsection{Algorithmic Problems}

Our algorithmic problems largely originate from programming contests (\eg Codeforces, AtCoder, ICPC, IOI) and classical CS problem settings (\eg knapsack), where tasks are solved under time and memory limits and are automatically judged. However, most contest problems admit a single optimal solution and are scored with a binary pass-or-fail system, which does not meet our requirement for open-ended problems without a known optimum. To address this mismatch, we introduce a structured curation pipeline, \emph{i.e.}, Proposal, Implementation, and Review, to construct genuinely variants.

\vspace{-12pt}
\paragraph{Proposal.} This stage is led by experts with qualifications equivalent to ICPC World Finalists, who are responsible for submitting candidate problems, complete with links to their original sources and a description of intended modifications. The proposal is reviewed by experts against:
(i) openness and multiplicity of viable solutions; (ii) discriminative strength of the scoring scheme; (iii) clarity and completeness of the problem statement and data ranges.

\vspace{-12pt}
\paragraph{Implementation.} This stage:
(i) converts each problem into an open-ended variant by changing single-optimum objectives and introducing a partial scoring system;
(ii) standardizes input/output formats or, when appropriate, provides an interaction library;
(iii) implements a deterministic verifier to formally assess the validity of candidate solutions; and
(iv) delivers a human-written reference solution with a significant advantage over the best LLMs;
(v) provides the necessary configuration and test data for evaluation.

\vspace{-12pt}
\paragraph{Review.} 
After implementation, each problem is reviewed by another algorithmic‑problem expert to make sure that:
(1) the problem has no known optimal solution or a nearly optimal solution that leaves limited room for improvement;
(2) the scoring policy is objective and can meaningfully reflect progress;
(3) the human reference solution is significantly stronger than the best model's performance; and
(4) the evaluator is implemented correctly and the test data can comprehensively evaluate solutions.


In total, FrontierCS contains 107 high-quality algorithmic problems adapted from programming contests, including 29 optimization problems, 27 constructive problems and 51 interactive problems.
Each problem contains: a well-defined statement with formal constraints; generators for test cases; a deterministic verifier with a partial scoring system; an expert-authored reference solution; a suite of baseline implementations; a reproducible model evaluation harness with associated scripts and metrics; and thorough documentation on provenance and data decontamination.

\vspace{-2ex}
\paragraph{Scoring policy design.}
We primarily evaluate solutions using problem-specific quality metrics such as cost, density, or number of queries, rather than computational efficiency. This reflects the open-ended nature of our tasks, where the main challenge is designing effective algorithms or strategies. 
Like traditional competitive programming tasks, each problem includes strict time and memory limits, and any solution that exceeds these limits is considered invalid and receives no score. In this way, runtime and memory act as feasibility constraints rather than components of the scoring metric. Unless explicitly specified otherwise, runtime is never part of the scoring, and together with the resource limits, this ensures that higher scores reflect genuinely better strategies rather than solutions that rely on excessive compute.

\vspace{-8pt}
\paragraph{On contest subtasks vs. our scoring.}
In programming contests such as ICPC and IOI, \emph{subtasks} partition tests into groups with extra constraints; points are awarded only if \emph{all} cases in a subtask pass, yielding discrete, all-or-nothing partial credit. 
Moreover, these subtasks are typically not strongly related to the essence of the problem, such as simplifying the problem's constraints or addressing the input data size.
Our benchmark instead uses task-specific, objective metrics (\eg density, cost, number of queries) with continuous or piecewise-continuous mappings to scores, measured relative to a trivial baseline and a strong human reference. This design rewards incremental improvements and enables fair comparison even when no single optimal solution is known.

\subsection{Research Problems}


\paragraph{Proposal.}
For FrontierCS research problems, we ask CS PhD students to set problems based on their unsolved research questions and to implement the evaluation container. 
Some of the research problems we select are drawn from recent AI-Driven Research for Systems (ADRS) work~\cite{cheng2025barbarians}, including multi-region spot-instance scheduling, cloud transfer path optimization, and LLM-guided SQL query reordering.
Similarly to the algorithmic problems, we use a process of \emph{proposal} and \emph{implementation}, with \emph{review} stages in between. 

\vspace{-12pt}
\paragraph{Implementation.} 
\begin{figure}[t]
    \centering
    \includegraphics[width=0.9\linewidth]{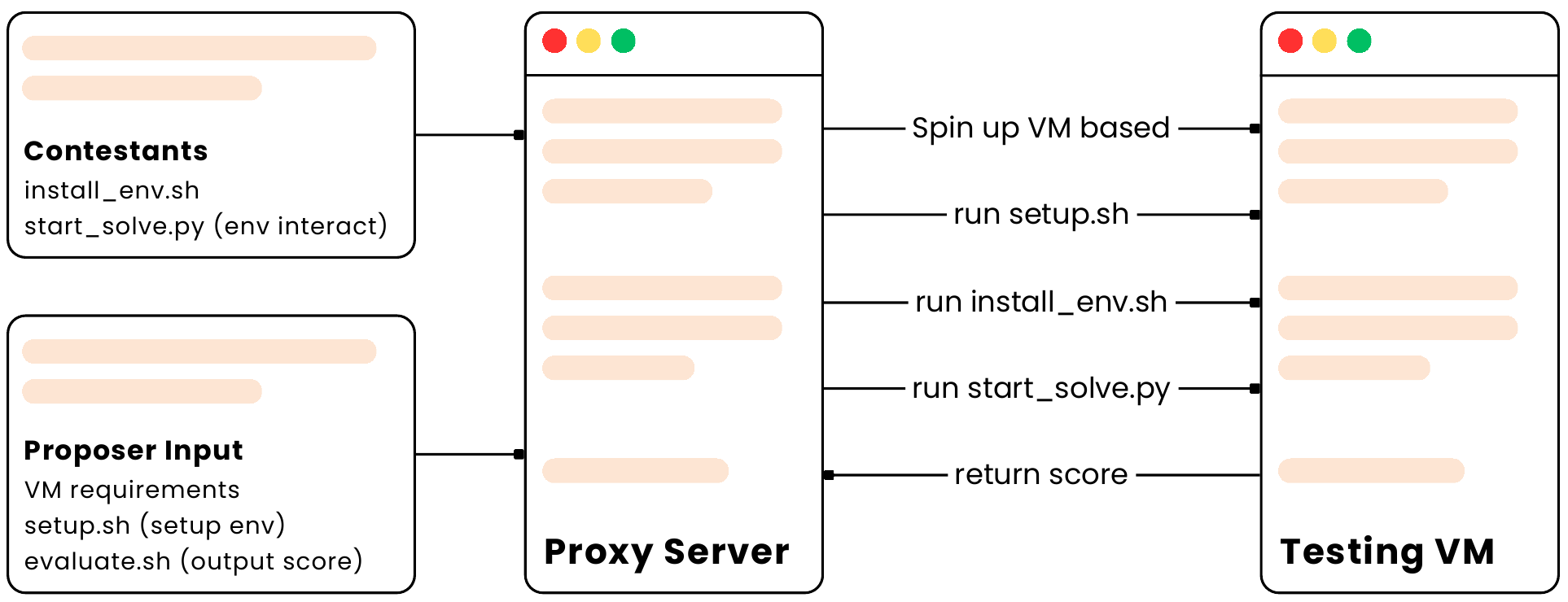}
    \caption{\textbf{Evaluation pipeline} of FrontierCS research problem using SkyPilot~\cite{skypilot}}
    \label{fig:research_problem_evaluation}
\end{figure}

Unlike problems sourced from programming contests that can be judged in a unified framework, each research problem needs a specialized environment. 
This stage ensures full reproducibility and objective evaluation. Candidate research problems contains resources, set\_up\_env.sh, evaluate.sh, and README file.

The \texttt{README} describes the problem statement, VM or Docker requirements, the data layout prepared by \texttt{set\_up\_env.sh}, and the exact input/output contract of the participant's solution. The environment can be utilized by the LLM or agent to create its solution for the problem. Examples include a training set for image classification problems.
Last but not least, the problem must support automated scoring via \texttt{evaluate.sh}, which gives a standardized score between 0 and 100. The evaluation has to be deterministic and must not use an LLM as a judge.
An overview of the architecture of the research evaluation platform is shown in Figure \ref{fig:research_problem_evaluation}.

\vspace{-2ex}
\paragraph{Infrastructure.}
To facilitate extensive experimentation, we integrate SkyPilot~\cite{skypilot} to manage the compute infrastructure. 
This abstraction allows the evaluation to scale up seamlessly from a single node to distributed cloud clusters, handling the heterogeneous hardware requirements of our research track. 
Crucially, it optimizes for economic viability by leveraging spot instances and region arbitrage, enabling researchers to evaluate agents on the full benchmark at minimal costs.

\vspace{-12pt}
\paragraph{Review.} 
Each submission is reviewed by CS researchers according to the following criteria:
(i) the problem should not have a single optimal solution;
(ii) partial scoring system should meaningfully reflect progress;
(iii) all scripts must run unattended in a fresh VM;
(iv) the environment must be deterministic, isolated, and free from external dependencies. 

\vspace{-2ex}
\paragraph{Scoring policy design.}
Unlike algorithmic problems, which are scored primarily on solution quality, research problems often involve multiple objectives such as accuracy, latency, memory usage, and cost due to their real-world nature. For example, a task may require designing a vector database index that minimizes query latency while maintaining at least 95\% accuracy (\probref{7} in Section~\ref{sec:example}). Nevertheless, we still impose strict resource limits for each problem to prevent excessive computation, and any solution exceeding the limit is considered invalid and receives no score.

\vspace{-12pt}
\paragraph{Variants.}
For each research problem, we provide multiple variants with different resource constraints and objective targets to reflect real-world research scenarios.
For example, a variant may impose stricter memory limit but turning off accuracy requirement, changing the hardware setting from CPU to GPU, or adjusting the objective from latency minimization to throughput maximization.
Note that, we treat every variant as an independent problem in the reported totals, because differing resource constraints or objectives lead to distinct solution strategies and reflect how related tasks are separately evaluated in real computer science research practice.

In total, \benchmark contains 49 research problems across diverse domains such as symbolic regression, vector database design, and kernel optimization. Each accepted research problem includes:
(i) a detailed problem description and background motivation;
(ii) a reproducible environment with pinned dependencies or Docker image;
(iii) a deterministic evaluator with partial scoring and diagnostics;
(iv) an expert-authored reference solution;
(v) the most trivial baseline implementations for comparison.

\subsection{Update Policy}

A core design principle of FrontierCS is to enable measurable progress on open-ended tasks. Rather than using binary judgments, our evaluators provide a score of 0 -- 100 using human performance as the baseline, allowing for measurable feedback on incremental improvements. 

FrontierCS is designed to remain relevant as models improve by supporting three complementary forms of task evolution.

\textbf{(1) Adding new tasks.} When expanding the scope of the benchmark or introducing fundamentally new problem categories, we may add new tasks. This is the traditional route for extending a benchmark, but it is not the only mechanism FrontierCS provides.

\textbf{(2) Increasing the difficulty of existing tasks without changing the problem statement.}
A key feature of FrontierCS is the separation between the written problem description and the environment in which the task is instantiated. This decoupling allows us to preserve the original question text while making the task more challenging. Difficulty can be increased by tightening constraints (e.g., time or memory budgets, feasibility requirements), modifying workloads or datasets (e.g., larger or more adversarial instances), or adjusting optimization objectives (e.g., stricter accuracy or performance targets). These updates retain task continuity while ensuring that the benchmark keeps pace with advancing model capabilities.

\textbf{(3) Refining human reference solutions and evaluation thresholds.}
When models approach or surpass strong human baselines, we can refine the human reference solution, scoring rubric, or evaluation thresholds to provide finer-grained separation between capable models. This method increases difficulty without modifying either the task description or the environment, enabling continued measurable progress.

Taken together, these three mechanisms provide a flexible, scalable, and principled update policy that allows FrontierCS to evolve over time without repeatedly constructing entirely new problem descriptions, while still preserving comparability across versions.

%% file: tex/4_exp.tex
\section{Evaluation Results}

In this section, we report the metric design and the performance of the frontier models on~\benchmark.

\subsection{Setup}

In our evaluation, we tested 9 frontier models: GPT 5 Thinking~\cite{openai2025gpt5}, Gemini 2.5 Pro~\cite{gemini25pro2025}, Gemini 3.0 Pro~\cite{gemini3}, Grok 4~\cite{xai2025grok4}, Claude Opus 4.1~\cite{anthropic2025opus41}, Claude Sonnet 4.5~\cite{anthropic2025sonnet45}, Claude Opus 4.5~\cite{claude4.5opus}, DeepSeek V3.2 Thinking~\cite{deepseek-v32-2025}.
We compared their performance against human experts on both the algorithmic problems track and the research problems track.
We set time-out for each LLM request as 20 minutes.
The other configurations for the models are:
\begin{itemize}[itemsep=0.5pt, leftmargin=*]
    \item GPT-5 / GPT-5.1 Thinking, Grok 4, Claude Opus 4.5:
          set \texttt{reasoning\_effort} = \texttt{high}.
    \item Claude Opus 4.1 / Sonnet 4.5:
          set \texttt{max\_tokens} = 32{,}000 and \texttt{reasoning\_budget} = 20{,}000.
    \item Gemini 2.5 Pro / Gemini 3.0 Pro:
          set \texttt{thinking\_budget} = -1.
\end{itemize}

During evaluation, models are tested in a single-round setting: once they produce a solution, that output is final. They do not have the opportunity to run their code, inspect unit-test results, or iterate based on feedback. They also do not have access to a code editor, a Python environment, or any external tools. All inputs are provided purely as text, \emph{i.e.,} no diagrams or visual components are involved and each problem can be fully described in text alone.
We leave the evaluation of multi-round, agentic, tool-call-assisted, and evolve-style frameworks to future work.

\subsection{Metrics}

In \benchmark, no problem has a universally correct solution, so we cannot use accuracy as an evaluation metric.
Instead, we introduce a grading system that scores each solution relative to two reference points: the reference solution written by human experts and a trivial baseline solution. 
Meanwhile, for some problems that have nontrivial bounds on achievable performance, we also use those bounds for reference.
A detailed example can be found in Section~\ref{sec:example}. 


Specifically, for a test case, if a solution fails to reach the level of the trivial baseline, it receives a score of zero; if it surpasses the threshold defined by human expert reference solution or the nontrivial bound, it receives a full score. 
Scores between these two extremes are assigned based on the nature of the problem, ensuring that each task has its own fair and tailored grading system. 
In all evaluations, we do not use any agentic framework and require the model to directly output code. However, we still encourage submissions that utilize agentic frameworks for comparison and future extensions. 

We report both Score@1, Avg@5, and Score@5, as we observe that the model exhibits stochasticity and notable improvement across multiple attempts. 
Here, we define Score@k as the maximum score among the $k$ model trials, and Avg@k as the average score among those trials. 
We also report Pass@1 and Pass@5: the fraction of problems where the model achieves a non-zero score, \emph{i.e.,} produces a valid solution that exceeds the the most trivial baseline.

\subsection{Algorithmic Problems}

\begin{table}[h]
    \centering
    \begin{tabular}{lrrrrr}
    Model & Score@1 & ~Avg@5 & Score@5 & Pass@1 & Pass@5\\
    \toprule
    \textcolor{gray}{Human Experts} & \textcolor{gray}{$95.41$} & \textcolor{gray}{-} & \textcolor{gray}{-} & \textcolor{gray}{-} & \textcolor{gray}{-}\\
    Gemini 3.0 Pro    & \textbf{29.37} & \textbf{29.51} & \textbf{52.06} & \textbf{65.42}\% & \textbf{83.18}\% \\
    Claude Opus 4.5   & $14.95$ & $13.83$ & $25.83$  & $54.21$\% & $69.16$\% \\
    Grok 4            & $13.67$ & $13.25$ & $26.42$  & $14.95$\% & $46.73$\% \\
    GPT 5 Thinking    &  $10.87$ & $12.03$ & $24.35$  & $38.32$\% & 63.55\%\\ 
    GPT 5.1 Thinking  &  $11.80$ & $11.90$ & $21.59$  & $34.58$\% & 54.21\%\\
    DeepSeek 3.2 & 12.65 & 11.80 & 24.20 & 46.73\% & 69.16\% \\
    Gemini 2.5 Pro    &  12.53 & 11.13 & 24.60  & 38.32\% & $64.49$\%\\
    Claude Opus 4.1   & $6.91$ & $5.84$ & $13.31$  & $30.84$\% & $48.60$\% \\ 
    Claude Sonnet 4.5 & $5.84$ & $6.72$ & $15.66$  & $28.04$\% & $55.14$\%

    \end{tabular}
    \caption{\textbf{Algorithmic problems results.} We define {score@k} as the highest score achieved across $k$ runs, and {avg@k} as the average score over those $k$ runs. {Pass@k} denotes the proportion of runs whose results exceed the most trivial baseline, not exceed human expert solution.}
    \label{tab:cp}
\end{table}
As shown in Table~\ref{tab:cp}, frontier models still lag significantly behind human experts on the algorithmic problems track,
where the 8 frontier models achieve Score@1 of 10.87 (GPT 5 Thinking), 11.80 (GPT 5.1 Thinking), 12.53 (Gemini 2.5 Pro), \textbf{29.37} (Gemini 3.0 Pro), 13.67 (Grok 4), 14.95 (Claude Opus 4.5), 6.91 (Claude Opus 4.1), and 5.84 (Claude sonnet 4.5) respectively, compared to human experts' \textcolor{gray}{95.41}.
These results highlight the substantial gap that remains between current AI capabilities and human expertise in tackling open-ended algorithmic challenges.
As expected, we also observe that all 8 models demonstrate stronger performance when increasing sampling attempts from one to five, indicating that they can generate diverse solutions and benefit from multiple trials.
For instance, Score@5 improves over Score@1 by 6.40 -- 22.69 points across models. 
However, they still lag far behind human experts on challenging open-ended problems.

\subsection{Research Problems}
\begin{table}[h]
    \centering
    \begin{tabular}{lrrrrr}
    Model & Score@1 & ~Avg@5 & Score@5 & Pass@1 & Pass@5\\
    \toprule
    Claude Opus 4.5     & \textbf{29.40} & \textbf{32.31} & 44.47 & \textbf{57.14}\% & 75.51\% \\
    GPT 5.1 Thinking    & 28.39 & 29.32 & \textbf{47.21} & \textbf{57.14}\% & \textbf{83.67}\% \\
    Gemini 3.0 Pro      & 26.95 & 29.18 & 47.20 & \textbf{57.14}\% & 81.63\% \\
    GPT 5 Thinking      & 23.74 & 27.12 & 44.78 & 46.94\% & 77.55\% \\
    Claude Sonnet 4.5   & 24.65 & 25.64 & 38.34 & \textbf{57.14}\% & 71.43\% \\
    Gemini 2.5 Pro      & 21.04 & 21.50 & 39.07 & 40.82\% & 73.47\% \\
    Grok 4 Fast & 18.01 & 13.60 & 28.81 & 36.73\% & 55.10\% \\
    Claude Opus 4.1     & 13.10 & 13.23 & 30.14 & 32.65\% & 63.27\% \\
    DeepSeek 3.2 & 14.65 & 10.89 & 21.37 & 26.53\% & 42.86\% \\
    \end{tabular}
    \caption{\textbf{Research problems results.} We define each metric as the same as in Table~\ref{tab:cp}.}
    \label{tab:research}
\end{table}

For the research track, Table~\ref{tab:research} reports Score@1, Avg@5, Score@5, Pass@1, and Pass@5 across 8 frontier models.
Claude Opus 4.5 attains the best one-shot results (Score@1 29.40), while GPT 5.1 Thinking leads under multi-sample evaluation (Score@5 47.21).
Across models, additional sampling yields sizable gains, \emph{i.e.,} Score@5 improves over Score@1 by 6.72 -- 21.04 points.
Pass rates remain substantially higher than raw scores, indicating that models often produce workable but under-optimized solutions, mirroring the trend observed on the algorithmic track.

\section{Discussion}

\begin{figure}[h!]
    \centering
    \begin{minipage}[t]{0.98\linewidth}
        \centering
        \includegraphics[width=\linewidth]{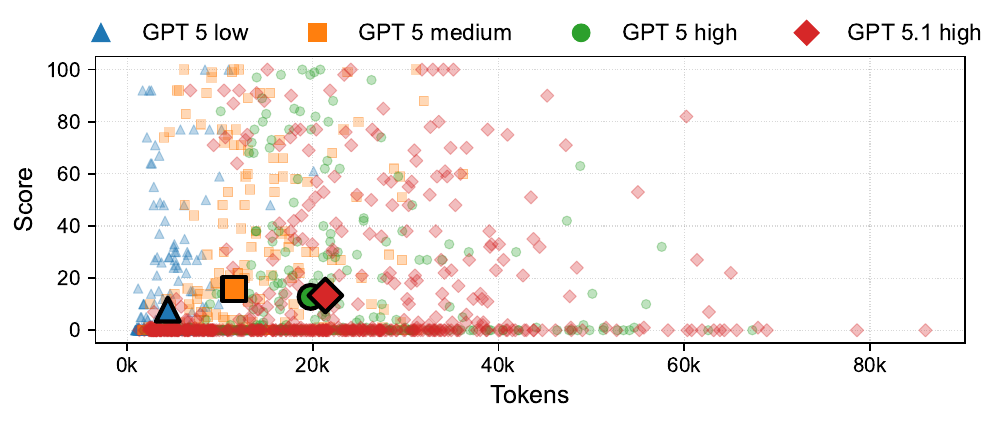}
    \end{minipage}\\
    \begin{minipage}[t]{\linewidth}
        \centering
        \small
        \begin{tabular}{lrr}
            Effort & Avg tokens & Avg score \\
            \midrule
            GPT 5 low    & 4,389 & 7.903 \\
            GPT 5 medium & 11,554 & \textbf{15.336} \\
            GPT 5 high   & 19,763 & 12.626 \\
            GPT 5.1 high & \textbf{20,402} & 12.508 \\
        \end{tabular}
    \end{minipage}
    \caption{\textbf{Reasoning tokens vs. Score}: We setting GPT 5 Thinking with different reasoning efforts. \textbf{Top}: scatter plot; Each point is one attempt (3 attempts per problem). Marks in the middle denote group averages. \textbf{Bottom}: average tokens and scores by reasoning effort.}
    \label{fig:reasoning_effort}
\end{figure}

\subsection{Improving Reasoning Effort Does Not Yield Further Gains}

We further analyze the relationship between reasoning effort and model performance on open-ended algorithmic problems.
As shown in Figure~\ref{fig:reasoning_effort}, we plot GPT 5 Thinking's score against the number of reasoning tokens it consumes per attempt by setting different \texttt{reasoning\_effort} levels, \emph{i.e.,} \texttt{low}, \texttt{medium}, and \texttt{high}.
As expected, we observe a clear positive correlation between reasoning effort when comparing low and medium reasoning levels.
However, increasing the reasoning effort from medium to high does not yield further gains; in fact, performance drops from 15.336 to 12.626, suggesting diminishing returns at higher reasoning budgets.
This indicates that while increased reasoning effort generally aids performance on open-ended problems, there may be an upper limit beyond which additional effort yields limited benefits for current LLMs.
Future work could explore more effective ways to leverage high reasoning effort for complex open-ended problem solving.

\subsection{Misleading Micro-Optimization Trap}  
During evaluation, we identify a recurring failure pattern in LLM behavior: the model often fixates on small, low-impact optimizations while overlooking the core algorithmic choices required for substantial performance gains. This is especially evident in Polyomino Packing (\probref{5} in Section \ref{sec:example}), which asks models to pack numerous polyominoes into a minimum-area rectangle and output a list of transformations for each piece. In practice, \text{GPT 5-Thinking} frequently adopts this transformation list as its internal data structure. Although memory-efficient and aligned with the final output format, this choice is a conceptual pitfall. Relying solely on transformation lists renders overlap detection and free-space search both cumbersome and error-prone. Consequently, the model produces invalid code in about 30\% of attempts, and in the remaining 70\% achieves only low scores, \emph{i.e.,} 20 -- 70 due to ineffective search strategies.

In contrast, a minimal prompt adjustment dramatically changes the outcome. Adding a single instruction like \emph{Please use a 2D array to maintain the rectangle state, and convert to the required format only at the end} reliably shifts the model toward a structurally sound internal representation. With this modification, the zero-score rate drops to about 10\%, and in nearly 80\% of cases the model successfully implements an efficient search strategy, achieving high scores in the 80 -- 85 range, consistently surpassing prior best solutions. This case study highlights a fundamental limitation of current LLMs: they do not inherently recognize which optimizations are algorithmically meaningful, often becoming trapped in superficially appealing but strategically irrelevant micro-optimizations.

\subsection{The Research–Engineering Dilemma of Claude}

In Table~\ref{tab:cp} and Table~\ref{tab:research}, we observe a clear disparity in Claude models' scores, whereas the other models perform similarly across both tracks. 
After examining its submissions, we find that Claude Sonnet 4.5 often produces workable solutions that generate basic outputs within resource limits and avoid compilation or runtime errors, yet these solutions frequently lack the optimization strategies needed for competitive performance. 
Consequently, most of its workable solutions fall below the trivial baseline in the algorithmic track and thus get \textbf{zero} score, leading to a lower overall score compared to models that generate fewer but higher-quality solutions.

In contrast, research problems require not only problem-specific optimization but also system-level reasoning and awareness of real research project environments; under these broader requirements, Claude Opus 4.5's workable solutions achieve relatively strong performance.
For example, in Symbolic Regression Problems (\probref{6} in Section~\ref{sec:example}), workable solutions must correctly invoke and configure the \texttt{PySR} package~\cite{cranmer2023interpretable} to search over expression spaces, tune evolutionary parameters, and validate candidate formulas against the provided datasets.
Although its optimization is limited, these workable solutions still earn around 50 points.

Interestingly, the high rate of workable solutions also aligns with Claude model's state-of-the-art performance on SWE-bench Verified~\cite{openai2024swebenchverified}.
This suggests that models tuned for closed-form software engineering tasks may still struggle to produce high-performance solutions for open-ended algorithmic problems. 
Overall, these results highlight that \textbf{(1)} merely producing workable solutions is insufficient for open-ended tasks, especially for algorithmic problems, and \textbf{(2)} success on research problems requires both engineering skills (\emph{e.g.,} knowing how to use existing research tools) to produce workable solutions and the ability to effectively optimize them.

%% file: tex/4_examples.tex
\clearpage
\section{Example Problems}
\label{sec:example}


\paragraph{Example (1)}
The first problem is adapted from the International Olympiad in Informatics (IOI) 2025, where FrontierCS's adaptation uses an open-ended grading policy to encourage more compact solutions, \emph{i.e.,} smaller grid sizes. The problem statement is as follows:

\begin{sampleproblem}{Problem 1: World Map}
\problabel{prob:1}
	\pbsection{Problem Description}
	You are given $N$ countries and a set $E$ of $M$ unordered pairs indicating which countries must be adjacent. Construct a $K\times K$ grid (for some integer $K$) that assigns each cell a country label from $\{1,\dots,N\}$ such that: 
	\begin{enumerate}
		\item For every $\{a,b\}\in E$, there exists at least one pair of orthogonally adjacent cells labeled $a$ and $b$;
		\item Whenever two orthogonally adjacent cells have distinct labels $a$ and $b$, then $\{a,b\}\in E$.
	\end{enumerate}
	Adjacency is by shared edge only (no diagonals). The goal is to minimize $K$.

	\pbsection{Grade Policy}
	If the output grid does not satisfy the adjacency requirements, the score is 0. Otherwise, if the output grid has size $K\times K$, let $R = \frac{K}{N}$, and let $R'$ be the same ratio from human expert solutions.
	The score is computed as 
	\[
	\text{score} = 100 \times \text{clamp}\left(\frac{6 - R}{6 - R'}, 0, 1\right)
	\]
\end{sampleproblem}

For this problem, the only known solution achieves $R' = 1.5$ across all possible inputs from an IOI 2025 submission. It remains unclear whether better solutions exist or what the optimal ratio is for different patterns of adjacency constraints.
Nevertheless, solutions can still be objectively graded by their achieved ratios, and solution validity can be easily verified by checking each adjacency requirement.

\begin{figure}[h!]
	\centering
    \resizebox{0.95\linewidth}{!}{
	\begin{subfigure}[t]{0.3\textwidth}
		\centering
		\begin{minipage}[t][3cm][c]{\linewidth}
			\centering
			\begin{tikzpicture}[scale=0.75,
				every node/.style={circle, draw=brandblue, fill=brandblue!5, inner sep=1.5pt, minimum size=16pt},
				edge/.style={line width=0.8pt, brandblue}
			]
				\def\r{1.6}
				\node (1) at (90: \r) {1};
				\node (2) at (18: \r) {2};
				\node (3) at (-54: \r) {3};
				\node (4) at (234: \r) {4};
				\node (5) at (162: \r) {5};
				\draw[edge] (1) -- (2);
				\draw[edge] (1) -- (3);
				\draw[edge] (1) -- (4);
				\draw[edge] (1) -- (5);
				\draw[edge] (2) -- (3);
				\draw[edge] (2) -- (4);
			\end{tikzpicture}
		\end{minipage}
		\caption{Adjacency graph $E$.}
		\label{fig:ex1-input}
	\end{subfigure}\hfill
	\begin{subfigure}[t]{0.3\textwidth}
		\centering
		\begin{minipage}[t][3cm][c]{\linewidth}
			\centering
			\includegraphics[height=3cm, keepaspectratio]{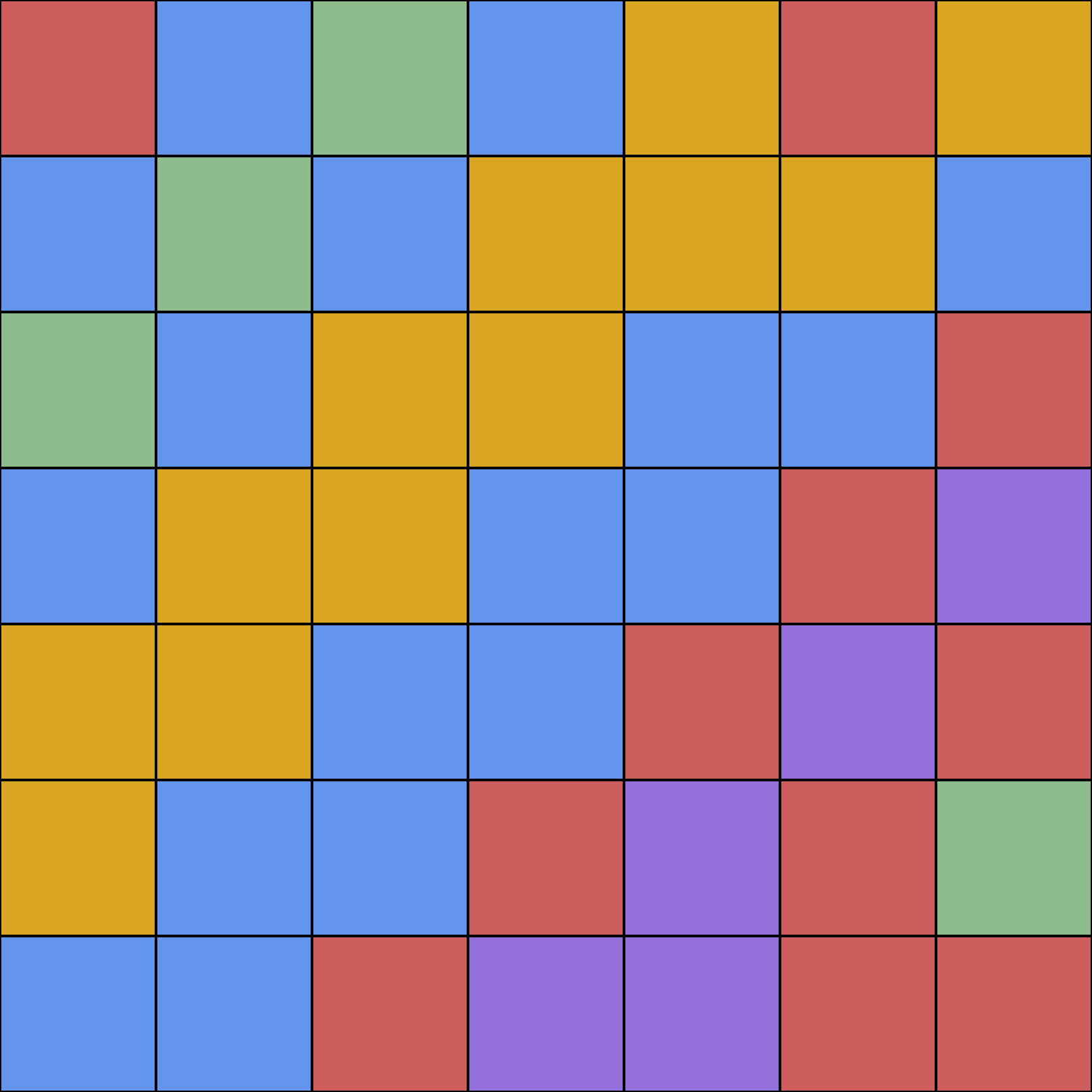}
		\end{minipage}
		\caption{Human expert ($K=7$).}
		\label{fig:ex1-human}
	\end{subfigure}\hfill
	\begin{subfigure}[t]{0.3\textwidth}
		\centering
		\begin{minipage}[t][3cm][c]{\linewidth}
			\centering
			\includegraphics[height=3cm, keepaspectratio]{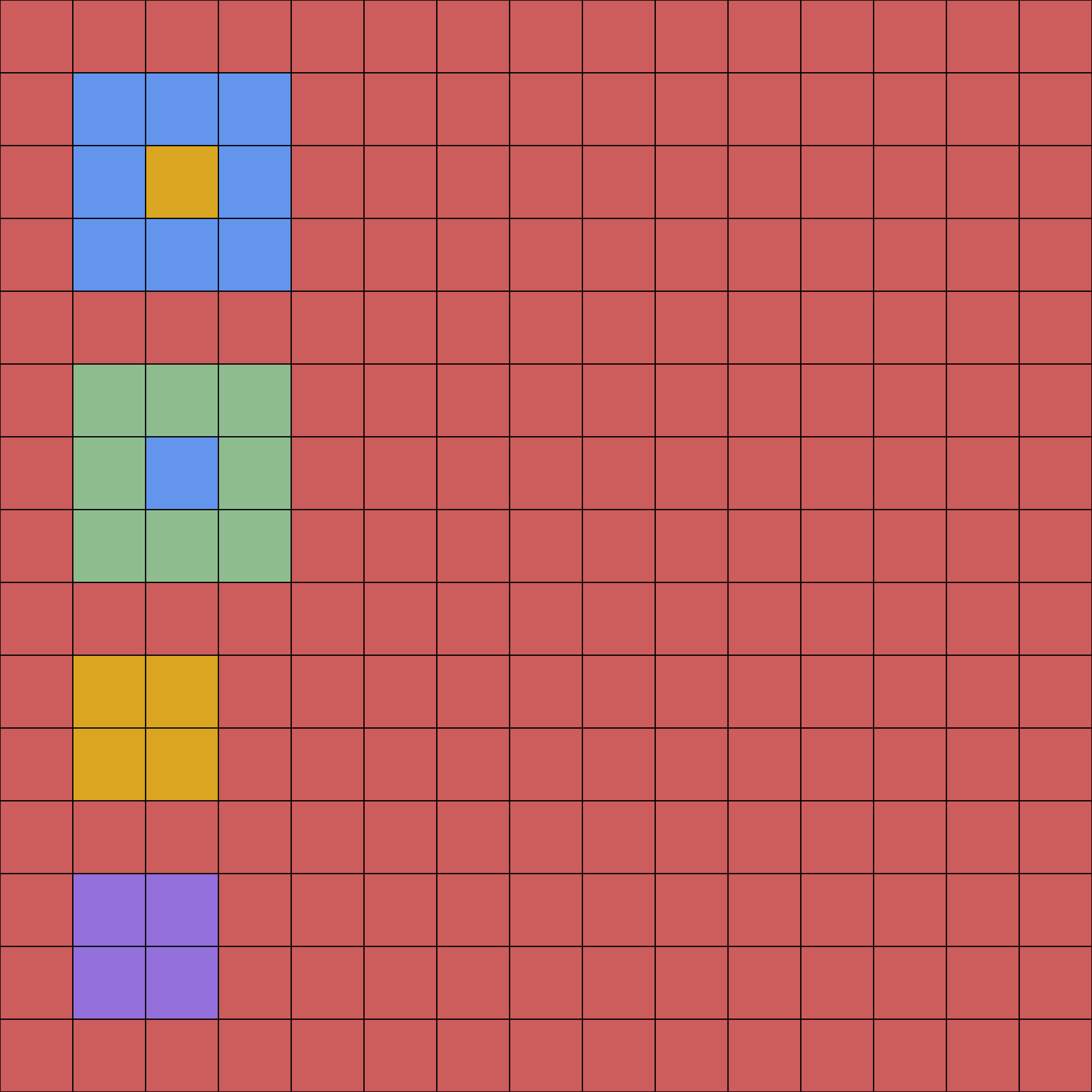}
		\end{minipage}
		\caption{GPT 5 ($K=15$).}
		\label{fig:ex1-llm}
	\end{subfigure}
    }
	\caption{Problem 1: Input–output illustration. (a) The input adjacency graph $E$ for $N=5$; (b) and (c) show valid outputs from a human expert's algorithm and a GPT 5 generated algorithm with respective grid sizes $K$.}
	\label{fig:problem1-examples}
\end{figure}
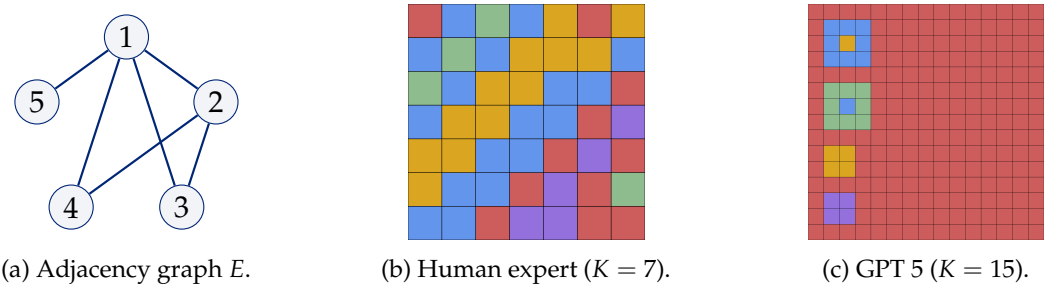

Here, we provide visualized solutions generated by GPT 5 and human experts for comparison, as shown in Figure~\ref{fig:problem1-examples}. The illustrated input test case has $N = 5$ countries and $M = 6$ adjacency requirements. In this example, the LLM's solution results in a significantly larger grid size ($K = 15$) compared to the human expert's compact solution ($K = 7$).



\clearpage
\paragraph{Example (2)}
The second problem, a variant of the knapsack problem, is adapted from the 2024 ICPC North America Championship (NAC) NSA Challenge. The problem statement is as follows: 

\begin{sampleproblem}{Problem 2: Treasure Packing}
\problabel{prob:2}
	\pbsection{Problem Description}
	You are given $C = 12$ treasure categories. Each category $c$ has per‑item value $v_c$, mass $m_c$, volume $\ell_c$, and an upper bound $q_c$ on how many you may take. The bag has two capacities: mass $M$ and volume $L$. Choose nonnegative integer counts $x_c\le q_c$ to maximize the total value $\sum_c v_c x_c$ while keeping total mass $\sum_c m_c x_c$ at most $M$ and total volume $\sum_c \ell_c x_c$ at most $L$.

	Formally, you need to solve the following optimization problem:
	\begin{align*}
	&\max_{x\in\mathbb{Z}_{\ge 0}^C} \sum_{c=1}^C v_c x_c \\
		\text{s.t.}\quad &\sum_{c=1}^C m_c x_c \le M,\ \ \sum_{c=1}^C \ell_c	 x_c \le L,\ \ 0 \le x_c \le q_c\,.
	\end{align*}

	For all test cases, $1 \leq q_c \leq 10^4$, $1 \leq v_c \leq 10^6$, $1 \leq m_c \leq 20 \times 10^6$, and $1 \leq \ell_c \leq 25 \times 10^6$.

	\pbsection{Grade Policy}
	The solution must return a valid result within a 1-second time limit and a 1024 MB memory limit. If the output is invalid or exceeds resource limits, the score is 0. Otherwise, the score is computed as
	\[
		\text{score} = 100 \times \text{clamp}\left(\frac{\text{value} - \text{value}_{\text{ base}}}{\text{value}_{\text{ ref}} - \text{value}_{\text{ base}}}, 0, 1\right)
	\],
	where $\text{value}$ is the total value of the submitted solution, $\text{value}_{\text{ base}}$ is the value of a baseline solution provided, and $\text{value}_{\text{ ref}}$ is the value of a reference solution from human experts.

	Currently, the baseline solution is a greedy naive algorithm, and the reference solution is enhanced from the champion solution of the challenge.

\end{sampleproblem}

\vspace{2ex}
Note that within the time and memory limits, exact solutions are generally infeasible for large test cases. 
However, approximate solutions can still be objectively graded based on their total value, allowing for open-ended algorithm discovery and improvement.

In this problem, the strategies used by the human expert and the LLM (GPT 5) are as follows.
\begin{itemize}[leftmargin=*, itemsep=2pt]
	\item The human expert solution uses a combination of greedy and randomized selection that incrementally improves its result over the entire time limit. 
	\item The LLM solution tries something similar, using an initial greedy pass augmented by a branch-and-bound algorithm that recursively explores and then fixes item counts. The LLM obtains points on every test case for a total score of \emph{74} points, which is good but still far from the human score of \emph{100}. 
\end{itemize}

\clearpage
\paragraph{Example (3)}
FrontierCS also includes interactive problems that require adaptive querying. 
\begin{sampleproblem}{Problem 3: Permutation Guess}
\problabel{prob:3}
	\pbsection{Problem Description}
	There is a hidden permutation $\pi$ of $[n]=\{1,2,\dots,n\}$ with $n=1000$. Your task is to identify $\pi$ using the fewest possible queries. The problem is interactive: in each query, you submit a length-$n$ integer sequence $a=(a_1,\dots,a_n)$ with each $a_i\in[1..n]$. The judge returns a single integer
	\[
		f(a) \,=\, \bigl|\{\, 1 \leq i \leq n : a_i = \pi_i \,\}\bigr|,\\
	\]
	\emph{i.e.,} the number of positions where your sequence matches the hidden permutation exactly. 
	\pbsection{Interaction}
	\begin{itemize}[itemsep=0pt]
		\item Query: submit any integer sequence of length $n$, $a\in[1..n]^n$.
		\item Response: a deterministic integer $f(a)\in\{0,1,\dots,n\}$.
		\item Goal: after some number of queries, output a final sequence $\hat\pi$; the submission is accepted if $\hat\pi=\pi$, otherwise it is rejected.
	\end{itemize}
	\pbsection{Grade Policy}
	The score is based on the number of queries $Q$ used to correctly identify $\pi$. If the submission is rejected, the score is 0. Otherwise, the score is computed as
	\[
	\text{score} = 100 \times \text{clamp}\left(\frac{Q_{\text{base}} - Q}{Q_{\text{base}} - Q_{\text{ref}}}, 0, 1\right)
	\],
	where $Q_{\text{base}}$ is the number of queries used by a naive binary search strategy (approximately 10,000), and $Q_{\text{ref}}$ is the number of queries used by a divide-and-conquer solution from human experts (approximately 6,000).
\end{sampleproblem}

In this problem, the LLM strategy contains redundant steps and is highly inefficient compared to the human expert, \emph{e.g.,} 12 steps vs. 5 steps for the instance shown in Figure~\ref{fig:problem3-strategies}.
\begin{figure}[h]
	\centering
	\begin{subfigure}[t]{0.48\textwidth}
		\centering
		\begin{minipage}[t][6cm][t]{\linewidth}\vspace{0pt}
			\footnotesize
			\renewcommand{\arraystretch}{1.15}
			\setlength{\tabcolsep}{4pt}
			\begin{tabular}{@{}p{0.10\linewidth} p{0.25\linewidth} p{0.1\linewidth} p{0.44\linewidth}@{}}
				\textbf{Step} & \textbf{Query} & $f(a)$ & \textbf{Inference} \\
				\hline
				\noalign{\vskip 4pt}
				1 & \texttt{1122} & 2 & $1\in\{1,2\}$, $2\in\{3,4\}$ \\
				2 & \texttt{3344} & 0 & $3\in\{1,2\}$, $4\in\{3,4\}$ \\
				3 & \texttt{1411} & 2 & $1$ at pos $2$, $4$ at pos $3$ \\
				4 & \texttt{2223} & 0 & $2$ at pos $1$, $3$ at pos $4$ \\
				5 & submit &  NA & answer: $\pi = \texttt{1432}$ \\
			\end{tabular}
		\end{minipage}
		\caption{Human expert: divide-and-conquer.}
		\label{fig:perm-human}
	\end{subfigure}\hfill
	\begin{subfigure}[t]{0.48\textwidth}
		\centering
		\begin{minipage}[t][6cm][t]{\linewidth}\vspace{0pt}
			\footnotesize
			\renewcommand{\arraystretch}{1.15}
			\setlength{\tabcolsep}{4pt}
			\begin{tabular}{@{}p{0.10\linewidth} p{0.25\linewidth} p{0.1\linewidth} p{0.44\linewidth}@{}}
				\textbf{Step} & \textbf{Query} & $f(a)$ & \textbf{Inference} \\
				\hline
				\noalign{\vskip 4pt}
				1 & \texttt{2322} & 1 &  \\
				2 & \texttt{3222} & 1 &  \\
				3 & \texttt{2233} & 1 &  \\
				4 & \texttt{3233} & 1 &  \\
				5 & \texttt{3232} & 2 & $2\in\{2,4\}$, $3 \in \{1, 3\}$\\
                6 & \texttt{3233} & 1 &  \\
                7 & \texttt{3233} & 1 &  \\
                8 & \texttt{3332} & 2 & $2$ at pos $4$ \\
                9 & \texttt{3222} & 1 & $3$ not at pos $1$ \\
                10 & \texttt{2322} & 1 & $3$ not at pos $2$, at pos $3$ \\
                11 & \texttt{4222} & 1 & $4$ not at pos $1$, at pos $2$ \\
				12 & submit & NA & answer: $\pi = \texttt{1432}$ \\
			\end{tabular}
		\end{minipage}
		\caption{LLM-generated solution (GPT 5).}
		\label{fig:perm-llm}
		\end{subfigure}
	\caption{Problem 3, toy instance ($n = 4$; hidden $\pi=\texttt{1432}$): side-by-side strategies with their query transcripts.}
	\label{fig:problem3-strategies}
\end{figure}

\clearpage
\paragraph{Example (4)}
Outside of more traditional competition problems, FrontierCS also includes classical open problems that require incremental optimization. 
\begin{sampleproblem}{Problem 4: Square Packing}
\problabel{prob:4}
	\pbsection{Problem Description}
	Given an integer $1 \leq n \leq 10{,}000$, you need to place $n$ unit squares inside an axis-aligned square container of side length $L$ such that:
	\begin{enumerate}[itemsep=0pt]
		\item Every unit square lies entirely inside the container, squares can be rotated by an arbitrary angle.
		\item Any pair of unit squares share no common interior points, corners or edges touching is allowed.
	\end{enumerate}
	Your goal is to give a valid output while minimizing $L$. 
	\pbsection{Grade Policy}
	The score is based on the size of the square container the solution code outputs. If the packing is invalid, the score is $0$. Otherwise, supposing the solution has size $L$, we define $L_B = \sqrt{n}$ as the absolute lower bound, and $L_0 = \lceil{\sqrt{n}}\rceil$ as the naive upper bound. We also define the size of a reference solution $s(n)$.

	The final score is computed as
	\[
	\text{Score} =
	\begin{cases}
	100 & \text{if } L = L_B \\
	95 + 5 \cdot \min\left(1.0, 1.1 \cdot \frac{s - L}{s - L_B}\right) & \text{if } L_B < L \le s \\
	94 \cdot \min\left(1.0, 1.1 \cdot \frac{L_0 - L}{L_0 - s}\right) + 1 & \text{if } s < L < L_0 \\
	0 & \text{if } L \ge L_0
	\end{cases}
	\]
\end{sampleproblem}

For this problem, best human solutions are available for $n \le 100$ (some proven optimal). We set the human benchmark to 95 points to leave space for better solutions; validity is easy to check.
For $n \leq 100$, we set $s(n)$ as the best human solution to date. 
For $n > 100$, we recursively define $s(n) = 2 \cdot s(\lceil n/4 \rceil)$. 
This is based on the divide-and-conquer strategy of packing four existing solutions of size $\lceil n/4 \rceil$ into one square container.

\begin{figure}[h]
	\centering
    \resizebox{1\linewidth}{!}{
	\begin{subfigure}[t]{0.5\textwidth}
		\centering
		\begin{minipage}[t][3cm][c]{\linewidth}
			\centering
			\includegraphics[height=2.78cm, keepaspectratio]{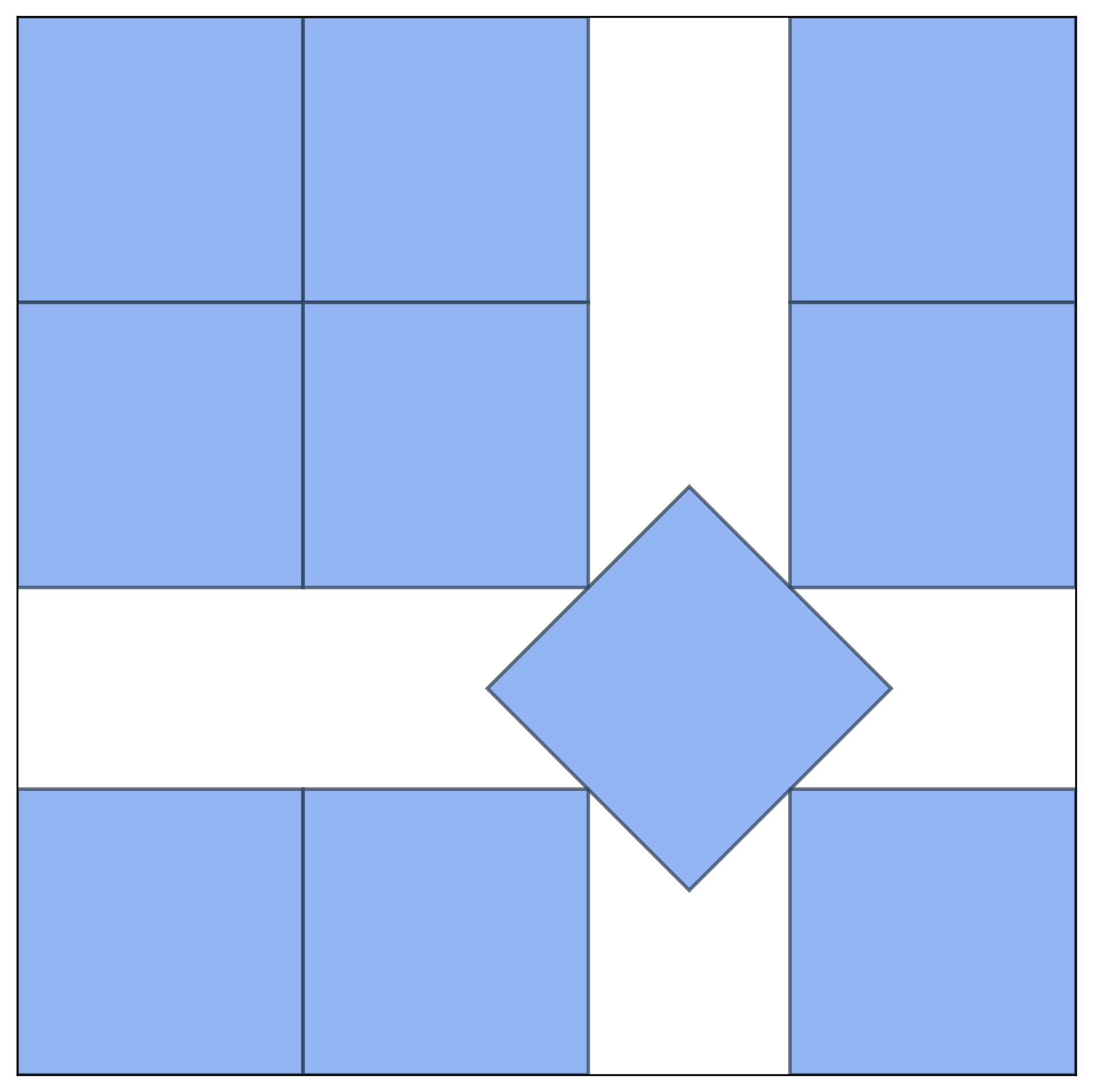}
		\end{minipage}
		\caption{Human expert ($L = 3.707$)}
		\label{fig:ex4-human}
	\end{subfigure}\hfill
	\begin{subfigure}[t]{0.5\textwidth}
		\centering
		\begin{minipage}[t][3cm][c]{\linewidth}
			\centering
			\includegraphics[height=3cm, keepaspectratio]{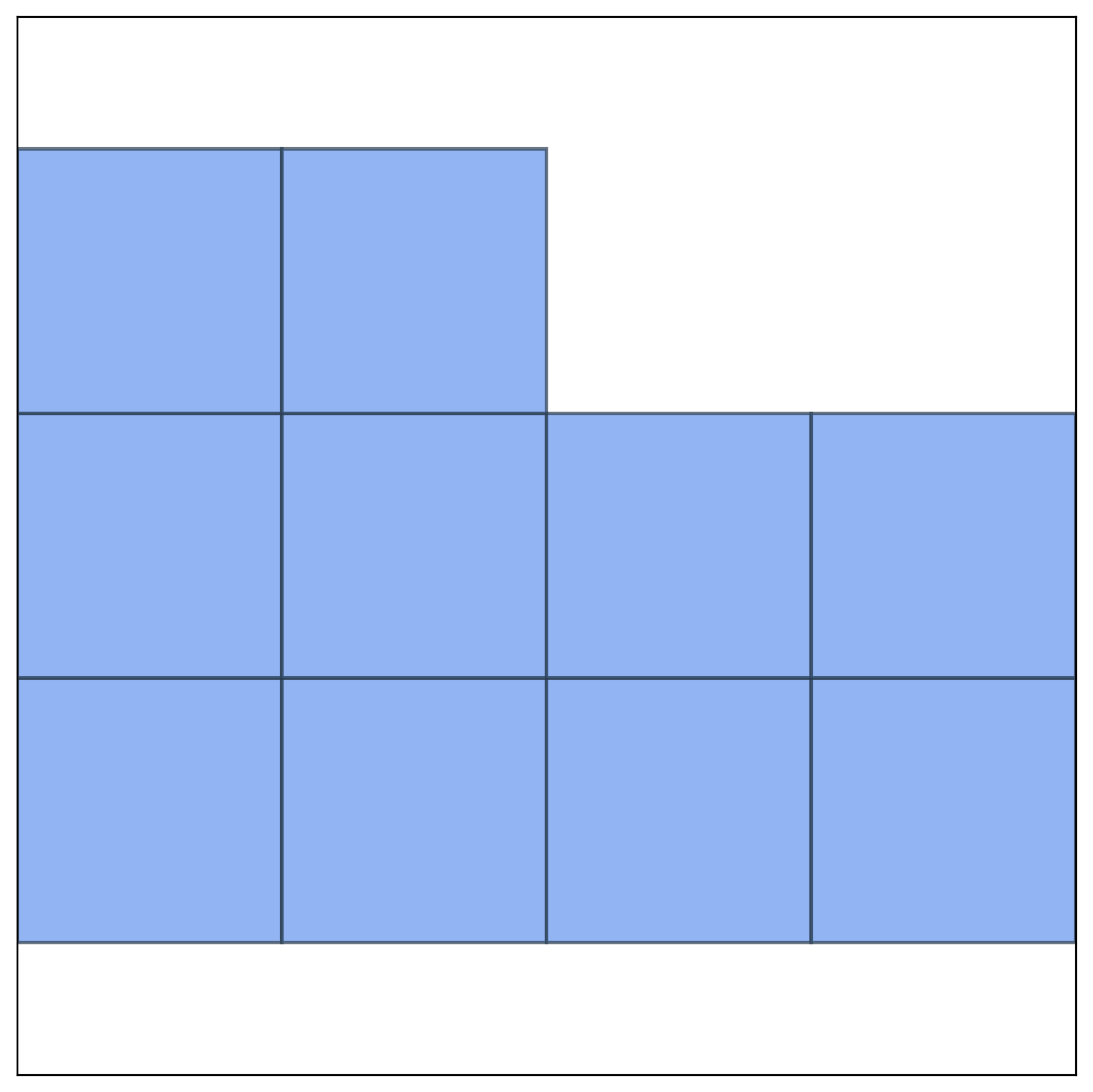}
		\end{minipage}
		\caption{Gemini 2.5 Pro ($L = 4$)}
		\label{fig:ex4-llm}
	\end{subfigure}
    }
	\caption{Problem 4: Shows valid outputs for $n = 10$ square packing from human expert and Gemini 2.5 Pro generated solution with respective square size $L$}
	\label{fig:problem4-examples}
	\vspace{-12pt}
\end{figure}

Figure~\ref{fig:problem4-examples} shows the $n=10$ case. The human solution achieves a minimal $L$ with a valid packing; the LLM uses a naive packing with $L=\lceil \sqrt{n} \rceil$. Note that for $n > 10$, most of the minimal $L$ is still open and unknown to human.

\clearpage
\paragraph{Example (5)}
\label{example-5}
A more challenging packing problem is Polyomino Packing (Figure~\ref{fig:teaser})

\begin{sampleproblem}{Problem 5: Polyomino Packing}
\problabel{prob:5}
	\pbsection{Problem Description}
	Given an integer $n$, the instance contains every distinct polyomino shape of size $s$ for all $s=1,\dots,n$. You need to place all these polyominoes into a $W\times H$ grid using the following allowed moves for each piece:
	\begin{itemize}[leftmargin=*, itemsep=0pt]
		\item integer translation $t_i\in\mathbb{Z}^2$,
		\item rotation by $0/90/180/270^\circ$ ($R_i\in\{0,1,2,3\}$),
		\item optional mirror across the $y$-axis ($F_i\in\{0,1\}$).
	\end{itemize}
	Valid placement:
	\begin{itemize}[leftmargin=*, itemsep=0pt]
		\item all transformed cells lie inside a $W\times H$ grid,
		\item no two pieces occupy the same grid cell (edge/corner touching is allowed).
	\end{itemize}
	Goal: minimize area $W\cdot H$ (equivalently, maximize density $\rho=$ packed cells $/(W\cdot H)$).

	\begin{center}
		\includegraphics[width=0.25\linewidth]{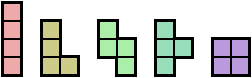}
		\captionof{figure}{Examples of polyominoes.}
	\end{center}

	\pbsection{Grade Policy}
	\begin{itemize}[leftmargin=*, itemsep=0pt]
		\item Validity: any out-of-bounds cell, overlap, or disallowed move gives score $0$.
		\item Scoring: let $\rho$ be your density. Score scales linearly from $0$ at the baseline density $\rho_{\text{base}}$ to $100$ at the reference density $\rho_{\text{ref}}$ (both provided per test set; $\rho_{\text{ref}}\le 1$).
	\end{itemize}
\end{sampleproblem}

\begin{figure}[h]
	\centering
	\vspace{1ex}
	\begin{subfigure}[t]{0.48\textwidth}
		\centering
		\begin{minipage}[t][3.5cm][c]{\linewidth}
			\centering
			\includegraphics[height=4cm, keepaspectratio]{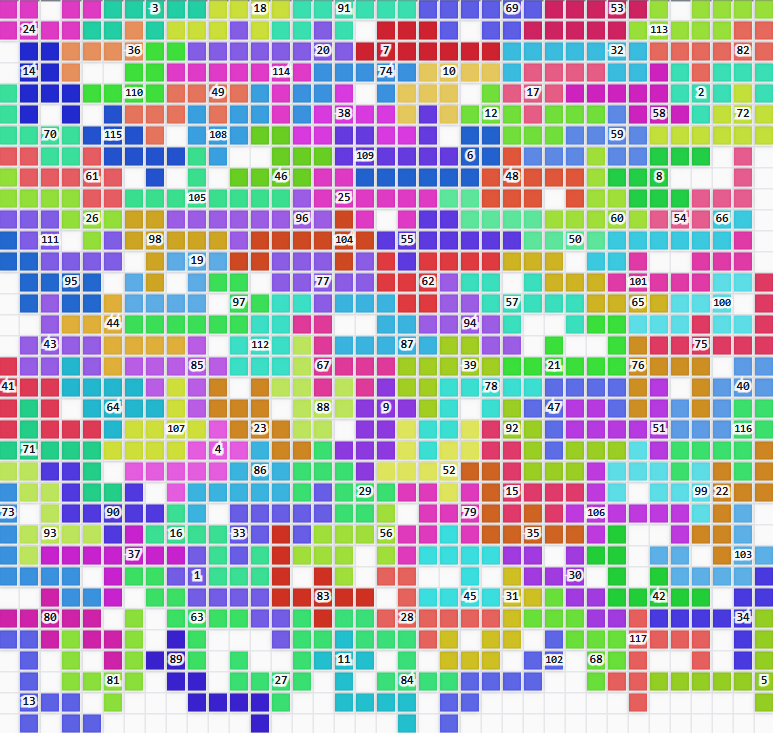}
		\end{minipage}
		\vspace{2ex}
		\caption{Human expert: 87\% density.}
		\label{fig:ex5-human}
	\end{subfigure}\hfill
	\begin{subfigure}[t]{0.48\textwidth}
		\centering
		\begin{minipage}[t][3.5cm][c]{\linewidth}
			\centering
			\includegraphics[height=4cm, keepaspectratio]{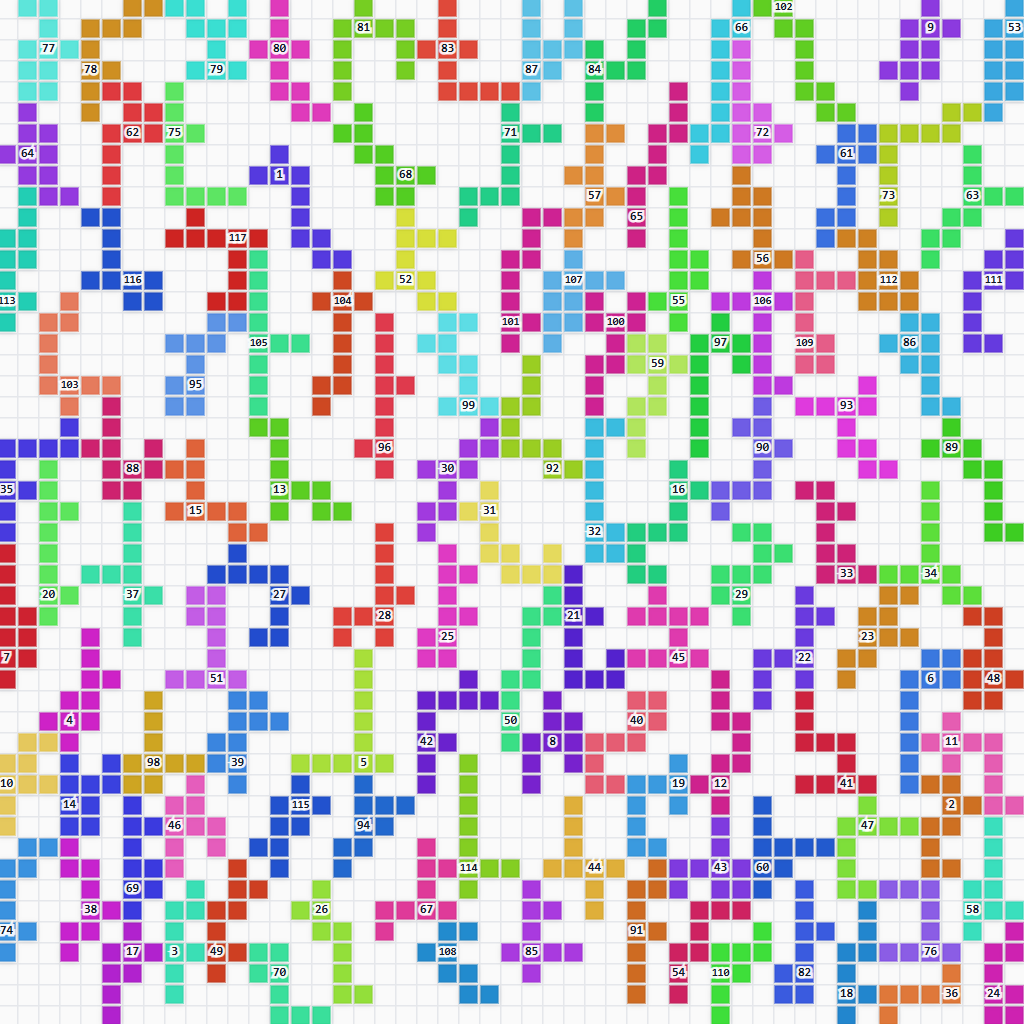}
		\end{minipage}
		\vspace{2ex}
		\caption{GPT 5-Thinking: 47\% density.}
		\label{fig:ex5-llm}
	\end{subfigure}
	\caption{Problem 5: Polyomino Packing — valid outputs from a human expert and GPT 5 for the same instance, illustrating density differences.}
	\label{fig:problem5-examples}
\end{figure}

Figure~\ref{fig:problem5-examples} shows a sample instance. The human expert achieves a noticeably tighter packing (higher density), while GPT 5 leaves more whitespace between pieces.

\clearpage
\paragraph{Example (6)}
FrontierCS also includes research problems such as symbolic regression.

\begin{figure}[h]
    \centering
    \begin{subfigure}[t]{0.33\textwidth}
        \centering
        \begin{minipage}[t][3.5cm][c]{\linewidth}
            \centering
            \includegraphics[height=4cm, keepaspectratio]{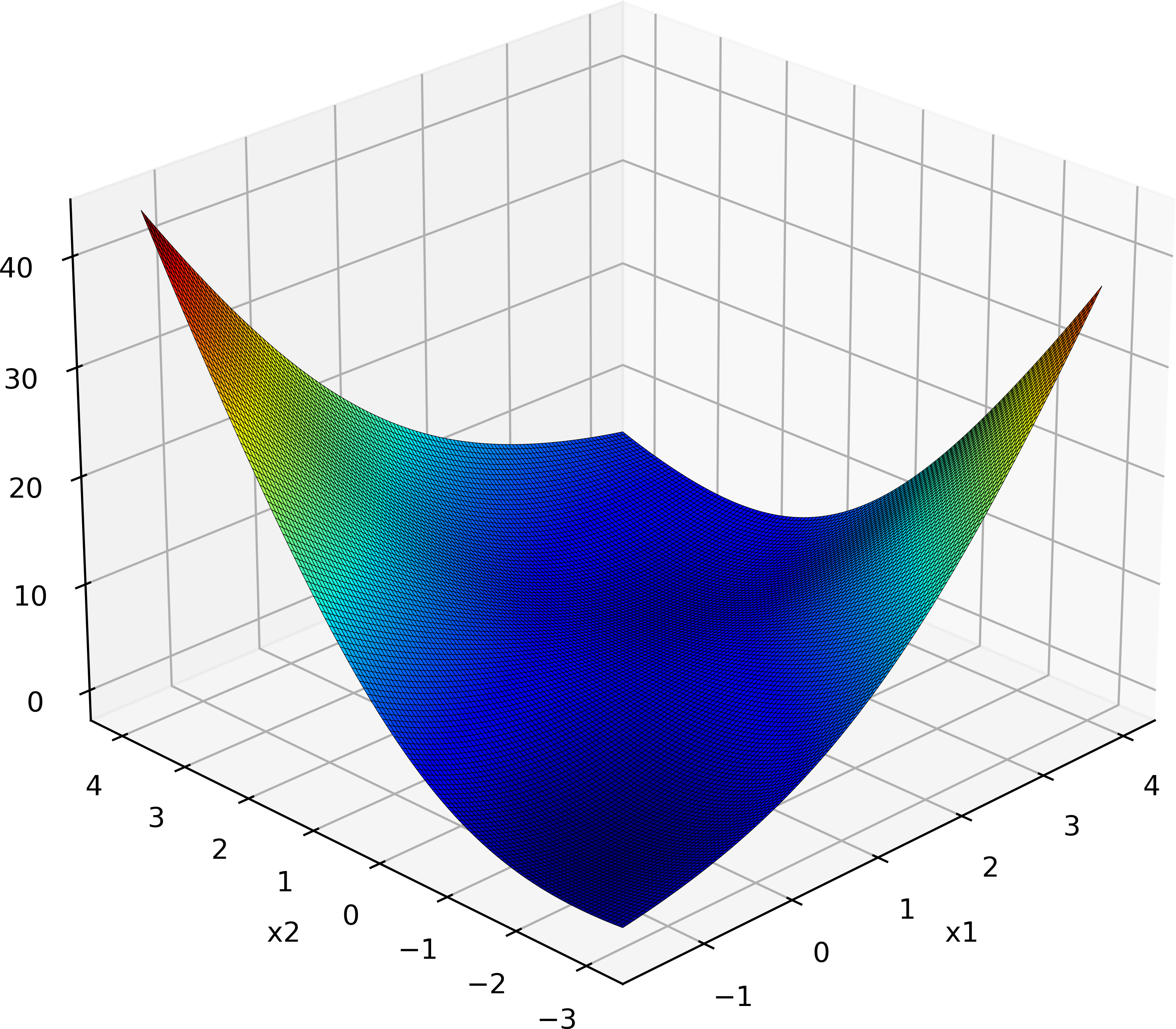}
        \end{minipage}
        \caption{Data plot.}
        \label{fig:ex6-truth}
    \end{subfigure}\hfill
    \begin{subfigure}[t]{0.33\textwidth}
        \centering
        \begin{minipage}[t][3.5cm][c]{\linewidth}
            \centering
            \includegraphics[height=4cm, keepaspectratio]{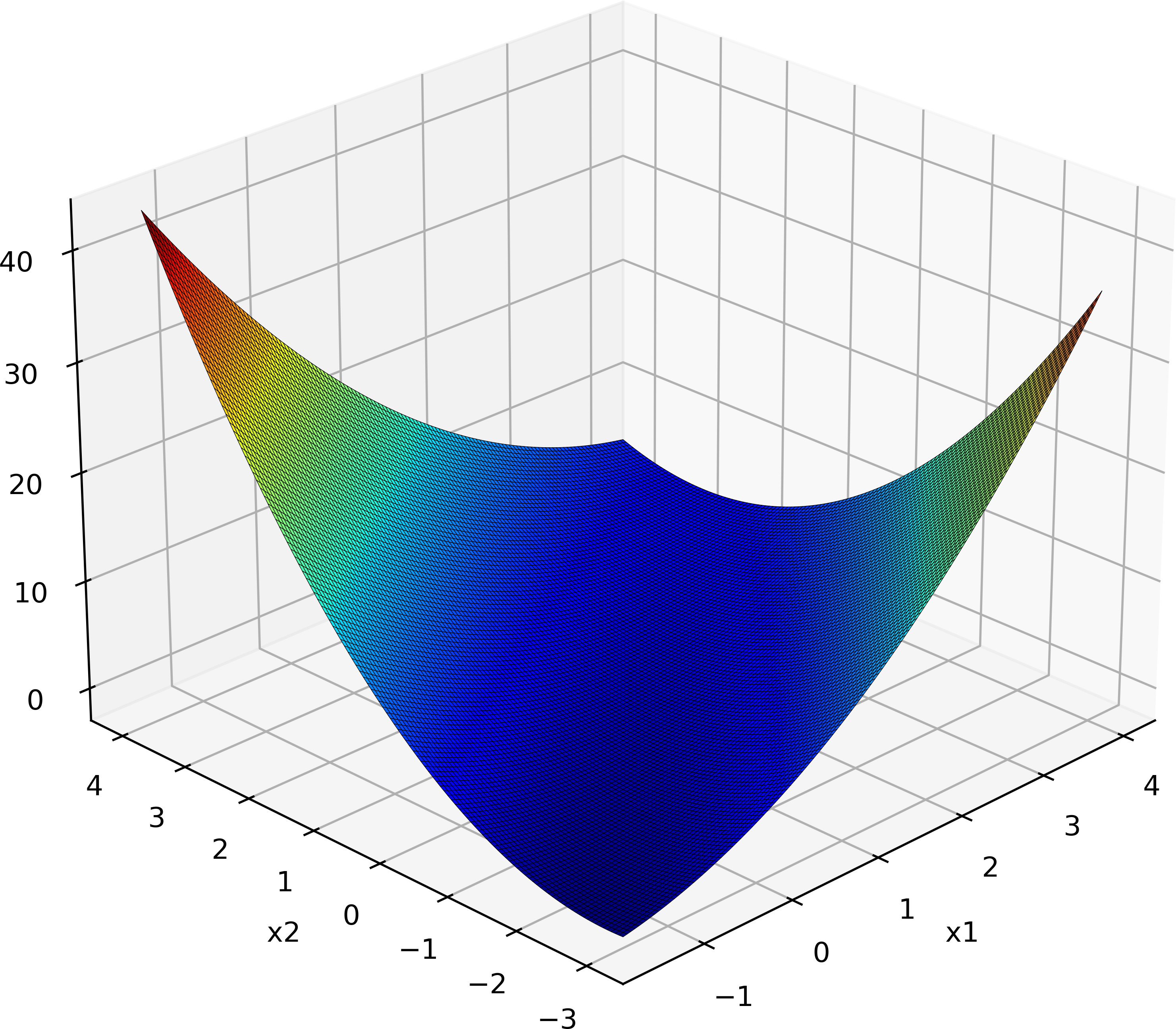}
        \end{minipage}
        \caption{Human expert (complexity 12).}
        \label{fig:ex6-human}
    \end{subfigure}\hfill
    \begin{subfigure}[t]{0.33\textwidth}
        \centering
        \begin{minipage}[t][3.5cm][c]{\linewidth}
            \centering
            \includegraphics[height=4cm, keepaspectratio]{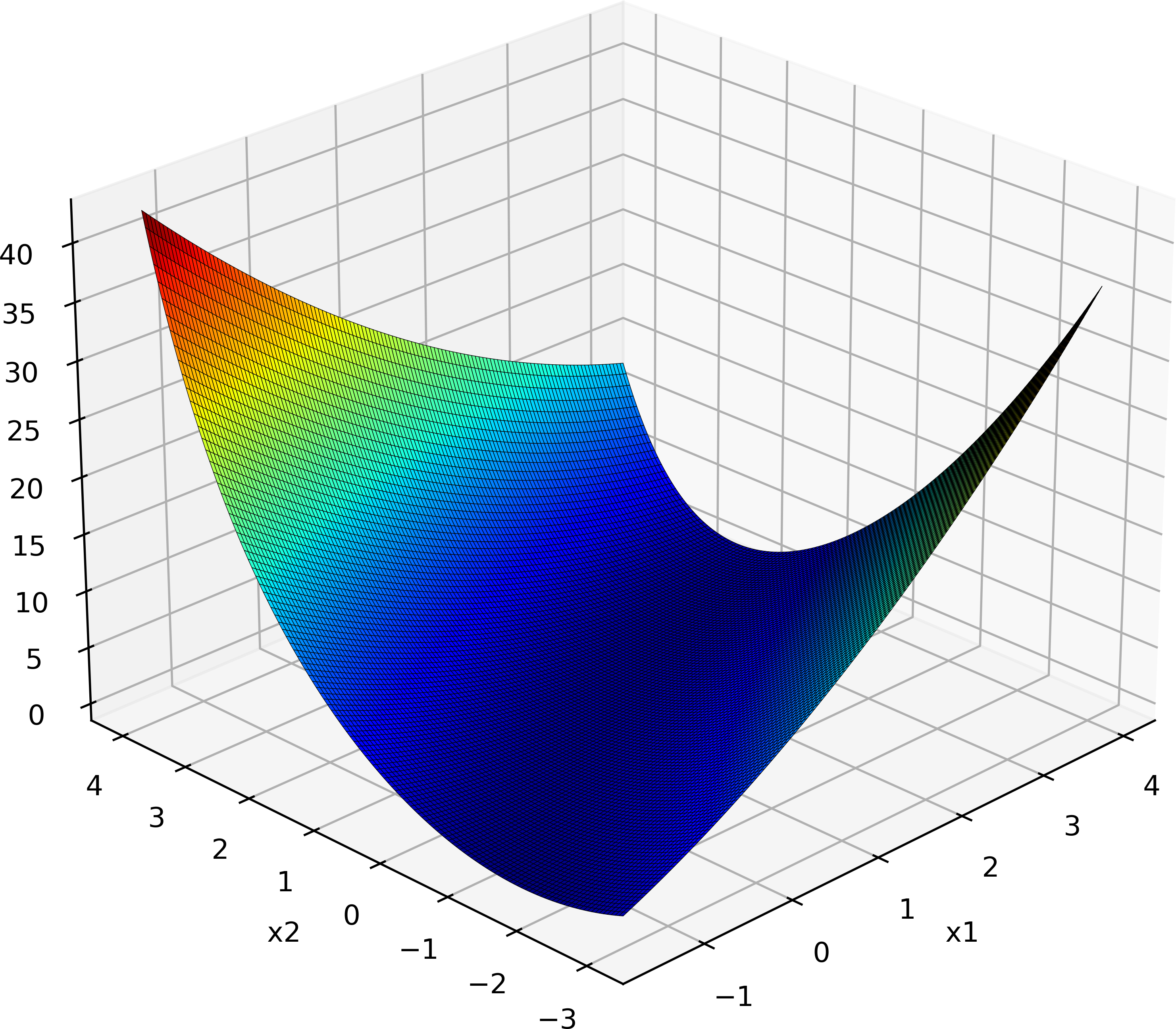}
        \end{minipage}
        \caption{GPT 5 (complexity 19).}
        \label{fig:ex6-llm}
    \end{subfigure}
    \caption{Problem 6: Symbolic Regression — data plot compare with functions discovered by a human expert and GPT 5, with noted expression complexities.}
    \label{fig:problem6-examples}
\end{figure}

\begin{sampleproblem}{Problem 6: Symbolic Regression}
\problabel{prob:6}
\pbsection{Problem Description}
Given a supervised dataset with features $x_1,\dots,x_d$ and target $y$, find a \emph{closed-form} expression $f(x_1,\dots,x_d)$ that predicts $y$ while keeping $f$ as simple as possible. The allowed grammar is:
\[
S \to S{+}S~|~S{-}S~|~S{\times}S~|~S{/}S~|~\exp(S)~|~\log(S)~|~\sin(S)~|~\cos(S)~|~(S)~|~x_j,\quad j\in\{1,\dots,d\}.
\]
Expressions must evaluate to finite real numbers on all rows of $X$ (no division by zero, $\log(\cdot)>0$, etc.); otherwise the submission is invalid and receives score $0$.

\pbsection{Complexity}
The expression complexity $\mathcal{C}$ is defined as
\[
\mathcal{C} \;=\; 2\times(\#\text{ binary ops}) \;+\; 1\times(\#\text{ unary ops}),
\]
which corresponds to the node count of the expression tree.

\pbsection{Grade Policy}
For feature matrix $X\in\mathbb{R}^{n\times d}$ and target $y\in\mathbb{R}^n$, let $\hat y=f(X)$ and
\[
\mathrm{MSE} \;=\; \frac{1}{n}\sum_{i=1}^n (y_i-\hat y_i)^2.
\]
Let $m_{\text{base}}$ be the MSE of the best linear predictor (OLS) on $(X,y)$, and $m_{\text{ref}}$ the MSE of the provided reference expression. The score is
\[
\text{Score} \;=\; 100\cdot \text{clamp}\!\left(\frac{m_{\text{base}}-\mathrm{MSE}}{m_{\text{base}}-m_{\text{ref}}},\,0,\,1\right)\cdot 0.99^{\max(\mathcal{C}-\mathcal{C}_{\text{ref}},\,0)},
\]
where $\mathcal{C}$ is the complexity of $f$ and $\mathcal{C}_{\text{ref}}$ that of the reference. If $m_{\text{base}}=m_{\text{ref}}$, set the score to $100$ when $\mathrm{MSE}\le m_{\text{ref}}$ and $0$ otherwise.

\end{sampleproblem}
Figure~\ref{fig:problem6-examples} shows the symbolic regression task for the McCormick function.
Figure~\ref{fig:ex6-truth} shows the direct plot of the data.
The human expert, with the help of symbolic regression tools, obtains the function shown in Figure~\ref{fig:ex6-human} with complexity 12, and GPT 5 produces the result in Figure~\ref{fig:ex6-llm} with complexity 19.

\clearpage
\paragraph{Example (7)}
Another problem included in FrontierCS is vector database system design.

\begin{sampleproblem}{Problem 7: Vector Database Design — Recall–Latency Tradeoff}
\problabel{prob:7}
	\pbsection{Problem Description}
	You need to build an approximate nearest-neighbor (ANN) index and evaluate on SIFT1M~\cite{jegou2010product} with
	1M base vectors (128D), 10K queries (128D), L2 metric, $k{=}1$.

	Report Recall@1 $r$ and average per-query latency $t$ in ms (search only).
	End-to-end run (build+search) must finish within 10 hours.

	\pbsection{Grade Policy}
	Validity requirements:
	\begin{enumerate}[leftmargin=*, itemsep=0pt]
		\item Implements the required API and runs within 10 hours;
		\item Returns finite distances and valid integer indices;
		\item Meets $r \ge r_{\min}$ and $t \le t_{\text{thr}}$; otherwise score $=0$.
	\end{enumerate}
	
	Scoring (if $t \le t_{\text{thr}}$):
	\[
		\text{Score} =
		100 \cdot \text{clamp}\!\Big(\frac{r - r_{\min}}{r_{\text{base}} - r_{\min}},\, 0,\, 1\Big).
	\]
	Here $r_{\min}$ is the minimum acceptable recall and $r_{\text{base}}$ is the reference recall of a strong human-designed index under the same latency budget $t_{\text{thr}}$.
\end{sampleproblem}

\begin{figure}[h]
	\centering
	\resizebox{0.62\linewidth}{!}{\input{fig/example7_vdb_design.tex}}
	\caption{Problem 7: Performance comparison of human expert and GPT 5-thinking in multiple VectorDB designing variants ($k = 1$).}
	 \label{fig:problem7-examples}
\end{figure}
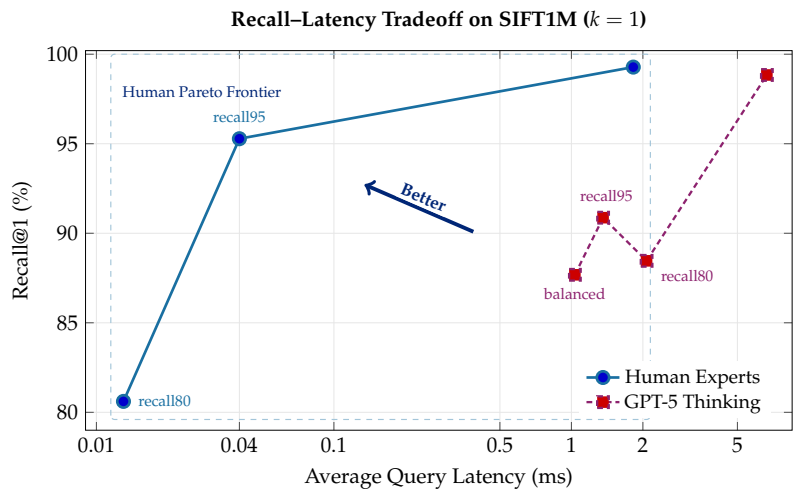

In practice, vector databases rely on approximate nearest neighbor (ANN) search, which inherently induces a latency-accuracy tradeoff: longer search budgets typically yield higher recall.
In this benchmark, we construct several task variants that explicitly reflect this property, \emph{e.g.,} \texttt{Recall80}, where the goal is to minimize latency while achieving at least 80\% recall.
Figure~\ref{fig:problem7-examples} compares GPT 5 Thinking with human experts across multiple such variants. As shown, GPT 5 Thinking substantially underperforms humans in these tasks.
Human results are obtained by tuning standard ANN index parameters in classic structures, \emph{e.g.,} adjusting \texttt{nprobe} in IVF or \texttt{efSearch} in HNSW.


\clearpage
\paragraph{Example (8)}
FrontierCS also includes research problems in cybersecurity.
\begin{sampleproblem}{Problem 8: Minimal PoC Generation}
\problabel{prob:8}
    \pbsection{Problem Description}
	Given a codebase and a target vulnerability description, generate a proof-of-concept (PoC) test to reproduce the vulnerability. Make the PoC as short as possible.
    \pbsection{Input}
    Each problem provides the following data:
    \begin{itemize}[itemsep=1pt]
		\item The entire pre-patch codebase of a real-world open-source project.
        \item The description of the target vulnerability.
	\end{itemize}
    Because the entire codebase may be too large for the current LLM context windows, we ask the LLM to generate a program that can process the codebase using the provided description. In addition, a solution may incorporate the LLM into an agentic workflow: it incrementally identifies relevant files, extracts the necessary inputs, and generates the target PoC.
    \pbsection{Grade Policy}
	The submitted PoC must successfully trigger the target vulnerability by meeting the following criteria:
    \begin{itemize}[itemsep=1pt]
		\item It triggers a sanitizer crash in the pre-patch codebase;
		\item It does not produce any sanitizer crash in the post-patch codebase;
		\item The correctness of the PoC is evaluated using the CyberGym~\cite{wang2025cybergym}.
	\end{itemize}
    Scoring:
    \[
	\text{Score} =
	\begin{cases}
		60 + 40 \times 2^{-\frac{L}{L_g}}, & \text{it triggers the target vulnerability}, \\
        0, & \text{it cannot trigger the target vulnerability},
	\end{cases}
	\]
    where $L$ denotes the length of the submitted PoC, and $L_g$ represents the length of the ground-truth PoC.
\end{sampleproblem}
Figure~\ref{fig:problem8-examples} shows valid PoC inputs that trigger a heap-use-after-free vulnerability in the PHP interpreter. A fuzzing program written by a human expert produces a 79-byte PoC, whereas GPT 5 generates a longer PoC of 577 bytes.

\begin{figure}[h]
	\centering
	\vspace{1ex}
	\begin{subfigure}[t]{0.48\textwidth}
		\centering
		\begin{minipage}[t][3.5cm][c]{\linewidth}
			\centering
\small
\begin{verbatim}
ff24 2201 6179 6972 6261 656c 2022 3c3d
703f 7068 6620 726f 2820 8024 ffff 3bff
2424 2e68 243d 6824 3d2f 6924 322d 2e30
3b30 6924 2b2b 2e29 3730 242d dfdf dfe0
695f 3f31 0a3e 9e3f 0a24 3053 0a0a 003f
\end{verbatim}
		\end{minipage}
		\caption{Human expert (79 bytes, hexadecimal dump).}
		\label{fig:ex8-human}
	\end{subfigure}\hfill
	\begin{subfigure}[t]{0.48\textwidth}
		\centering
		\begin{minipage}[t][3.5cm][c]{\linewidth}
			\centering
\small
\begin{verbatim}
3f3c 6870 0a70 2f2f 4620 726f 6563 6520
7261 796c 642d 7365 7274 6375 6974 6e6f
5520 4641 7620 6169 6320 6d6f 6f70 6e75
2064 6964 6976 6564 612d 7373 6769 206e
7962 7a20 7265 2e6f 2f0a 202f 484c 2053
...
200a 2020 6520 6863 206f 7573 7362 7274
2428 2c62 3020 202c 2931 0a3b 0a7d 3e3f
000a
\end{verbatim}
		\end{minipage}
		\caption{GPT 5 (577 bytes, hexadecimal dump).}
		\label{fig:ex8-llm}
	\end{subfigure}
	\caption{Problem 8: Minimal PoC Generation - valid PoC of a heap-use-after-free vulnerability.}
	\label{fig:problem8-examples}
\end{figure}

\clearpage
\paragraph{Example (9)}
FrontierCS also includes kernel code optimization problems that mirror real-world systems workflows. There are two types of problems: (A) rewrite a PyTorch fused implementation into a Triton kernel; (B) further optimize the Triton kernel using warp specialization.

\begin{sampleproblem}{Problem 9: Kernel Rewrite \& Warp Specialization for GDPA}
\problabel{prob:9}
  \pbsection{Problem Description}
  Given query ($Q$), key ($K$), value ($V$) tensors and per-element gates ($G_Q, G_K$), compute \emph{gated dot-product attention}
  \[
  \begin{aligned}
	d &= \text{head dimension},\qquad \alpha = 1/\sqrt{d},\\
	\tilde Q &= Q \odot \sigma(G_Q),\qquad \tilde K = K \odot \sigma(G_K),\\
	S &= \alpha\, \tilde Q\, \tilde K^\top,\qquad
	P = \operatorname{softmax}(S)\ \text{(row-wise over keys)},\\
	O &= P\, V.
  \end{aligned}
  \]
  Shapes follow Transformer convention. For batch size $B$, heads $H$, query length $M$, key/value length $N$, head dim $d$:
  \[
  Q,K,V\in\mathbb{R}^{B\times H\times(\cdot)\times d},\quad
  G_Q\in\mathbb{R}^{B\times H\times M\times d},\ 
  G_K\in\mathbb{R}^{B\times H\times N\times d},\ 
  O\in\mathbb{R}^{B\times H\times M\times d}.
  \]
  Inputs are \texttt{float16}; accumulations must be in \texttt{float32}; outputs are cast back to \texttt{float16} by default (checked by \texttt{rtol=1e-3, atol=5e-4}).

	A baseline PyTorch implementation is provided:
  \begin{lstlisting}[language=Python,basicstyle=\ttfamily\small]
import math, torch

def _pt_gdpa(Q, K, V, GQ, GK):
	scale = 1.0 / math.sqrt(Q.shape[-1])
	Qg = Q * torch.sigmoid(GQ)
	Kg = K * torch.sigmoid(GK)
	scores = torch.matmul(Qg, Kg.transpose(-1, -2)) * scale
	P = torch.softmax(scores, dim=-1)
	O = torch.matmul(P, V).to(torch.float16)
	return O
  \end{lstlisting}
  \vspace{1ex}

  \pbsection{(A) Triton Kernel Rewrite}
	Implement \textbf{\texttt{triton\_gdpa}} to compute GDPA using Triton (given Triton language and interface hints).
  \vspace{1ex}
  \pbsection{(B) Warp-Specialized Triton}
	Transform your Triton kernel into a \emph{warp-specialized} design using TLX (given the new DSL documentation).

  \vspace{1ex}
  \pbsection{Grade Policy}
	Performance is measured by time-saved fraction over the PyTorch baseline:
	\[
        \text{Score} = 100 \times (1-\frac{T_{\text{solution}}}{T_{\text{baseline}}})
	\]
	where $T_{\text{baseline}}$ and $T_{\text{solution}}$ are kernel runtimes across benchmark shapes.

\end{sampleproblem}

\begin{figure}[h]
	\centering
	\includegraphics[width=0.33\linewidth]{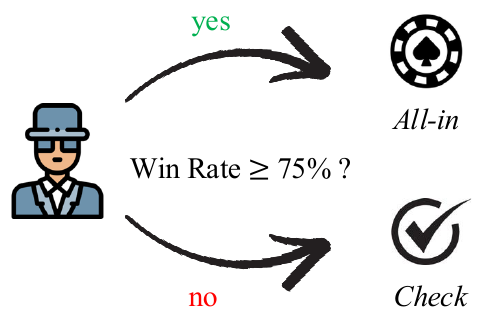}
	\caption{Problem 10: Human expert strategy.}
	\label{fig:problem10-examples}
\end{figure}

\clearpage
\paragraph{Example (10)}
FrontierCS also includes games and decision-making problems that require strategic planning. An example problem is as follows:
\begin{sampleproblem}{Problem 10: Poker Strategy Optimization}
\problabel{prob:10}
  \pbsection{Problem Description}
  You play heads-up Texas Hold'em against a fixed opponent strategy.

  \pbsection{Opponent policy}
  \begin{itemize}[leftmargin=*, itemsep=0pt]
	\item The opponent compares \texttt{Fold} EV (current chips $-\,100$) with \texttt{Call} EV estimated by Monte Carlo:
	  \begin{itemize}[leftmargin=1.5em, itemsep=0pt]
		\item Repeat 100 trials: uniformly permute the remaining unseen deck; fill all currently unknown cards (your hole cards and any unrevealed public cards) from the permutation.
		\item Assume that after the current decision both players only Check for the rest of the hand.
		\item Compute the opponent's payoff in that trial if they \texttt{Call} now.
	  \end{itemize}
	\item The opponent calls iff \texttt{Call} EV $>$ \texttt{Fold} EV; otherwise they Fold.
  \end{itemize}

  \pbsection{Implementation Task}
  Implement a routine decision making that, given your hole cards, revealed public cards, current pot and chip counts, and the current betting round, returns one of: \texttt{Check}, \texttt{Fold}, or \texttt{Raise($x$)} with $1\le x\le$ your remaining chips. The judge will drive many independent hands, invoking your decision at each betting round and applying the rules and opponent policy above.

  \pbsection{Grade Policy}
  Let $\omega$ be the final average profit per hand. Your points are a piecewise-linear function of $\omega$:
\[
\text{score} =
\begin{cases}
0, & \omega \le 8.0,\\
13.3\,(\omega - 8), & 8.0 < \omega \le 11.0,\\
40 + 14\,(\omega - 11), & 11.0 < \omega \le 14.0,\\
82 + 3\,(\omega - 14), & 14.0 < \omega \le 20.0,\\
100, & \omega \ge 20.0.
\end{cases}
\]
\end{sampleproblem}
In this problem, GPT 5 and Gemini 2.5 pro got 25 and 36, respectively.
However, a simple strategy from human experts (shown in Figure~\ref{fig:problem10-examples}) can achieve 54 score, which also use the same Monte Carlo simulations to estimate the expected value of each action, but will always \texttt{All-in} when having $> 0.75$ winning probability in each turn.

%% file: fig/example7_vdb_design.tex
\begin{tikzpicture}
  \definecolor{humanColor}{RGB}{25,111,161}
  \definecolor{gptColor}{RGB}{141,34,110}


  \begin{axis}[
    width=0.85\linewidth,
    height=0.5\linewidth,
    xmode=log,
    log basis x=10,
    xlabel={Average Query Latency (ms)},
    ylabel={Recall@1 (\%)},
    xmin=0.009, xmax=8.5,
    ymin=79, ymax=100.2,
    xtick={0.01,0.04,0.1,0.5,1,2,5},
    xticklabels={0.01,0.04,0.1,0.5,1,2,5},
    ytick={80,85,90,95,100},
    scaled ticks=false,
    tick style={color=black!70},
    grid=major,
    grid style={gray!20},
    legend style={draw=none, font=\small},
    legend pos=south east,
    title={Recall--Latency Tradeoff on SIFT1M ($k=1$)},
    title style={font=\bfseries},
    clip=false,
  ]
    \addplot+[
      humanColor,
      line width=1.4pt,
      mark=*, mark size=3.2pt,
    ] coordinates {(0.013,80.61) (0.040,95.28) (1.822,99.28)};
    \addlegendentry{Human Experts}

    \addplot+[
      gptColor,
      line width=1.2pt,
      dash pattern=on 3pt off 2pt,
      mark=square*, mark size=3.0pt,
    ] coordinates {(1.0344,87.68) (1.3596,90.87) (2.0728,88.46) (6.6537,98.84)};
    \addlegendentry{GPT-5 Thinking}

  \node[font=\scriptsize, anchor=north west, humanColor] at (axis cs:0.014,81.4) {recall80};
  \node[font=\scriptsize, anchor=south, humanColor] at (axis cs:0.040,95.80) {recall95};

  \node[font=\scriptsize, anchor=north, gptColor] at (axis cs:1.0344,87.35) {balanced};
  \node[font=\scriptsize, anchor=south west, gptColor] at (axis cs:1.0146,91.4) {recall95};
  \node[font=\scriptsize, anchor=north, gptColor] at (axis cs:3.0728,88.35) {recall80};

    \draw[humanColor!40, rounded corners=2pt, dashed, line width=0.6pt]
      (axis cs:0.0115,79.6) rectangle (axis cs:2.15,100.0);
    \node[font=\scriptsize, brandblue, anchor=west] at (axis cs:0.012,97.9) {Human Pareto Frontier};

    \draw[->, line width=2.0pt, brandblue, shorten <=2pt, shorten >=2pt]
      (axis cs:0.4,90.0) -- (axis cs:0.13, 92.8)
      node[midway, sloped, above, fill=white, inner sep=2pt, rounded corners=1pt, font=\bfseries\footnotesize]{Better};
  \end{axis}
\end{tikzpicture}


%% file: tex/5_conclusion.tex
\section{Conclusion}

We introduced \benchmark, a comprehensive and diverse benchmark for open-ended computer science tasks where global optima are unknown but solutions remain deterministically verifiable and partially gradable. \benchmark provides expert reference solutions, an automated evaluator, and a fully reproducible pipeline, and adopts versioned difficulty scheduling to preserve discrimination as models improve. This fills the current gap where LLM-based efforts to tackle open-ended CS problems lack a comprehensive and systematic testbed. Our initial study shows that current LLMs remain brittle on open-ended optimization and system-level trade-offs, and that competence on closed-form coding tasks does not reliably translate into open-ended problem solving.

\section*{Acknowledgments} 
We thank Ce Jin, Mingrui Liu, Youliang Yuan, and Qingyu Shi for valuable discussions and feedback on the benchmark's design and evaluation. 